\documentclass[10pt]{article}
\usepackage[margin=1in]{geometry}
\usepackage{fontspec}
\newfontfamily\thaifont{Garuda.otf}[BoldFont=Garuda-Bold.otf, ItalicFont=Garuda-Oblique.otf,
  BoldItalicFont=Garuda-BoldOblique.otf, Scale=MatchLowercase, Script=Thai]
\XeTeXlinebreaklocale "th"
\newXeTeXintercharclass\thaiclass
\XeTeXinterchartoks 0 \thaiclass = {\begingroup\thaifont}
\XeTeXinterchartoks 4095 \thaiclass = {\begingroup\thaifont}
\XeTeXinterchartoks \thaiclass 0 = {\endgroup}
\XeTeXinterchartoks \thaiclass 4095 = {\endgroup}
\usepackage{amsmath,amssymb}
\usepackage{booktabs}
\usepackage{array}
\usepackage{tabularx}
\usepackage{microtype}
\usepackage{url}
\usepackage{tikz}
\usetikzlibrary{arrows.meta,positioning}
\usepackage{float}
\usepackage{enumitem}
\setlist{nosep}

\usepackage[hidelinks]{hyperref}
\usepackage[numbers,sort&compress]{natbib}
\usepackage{multicol}

\title{Single-Sample Decoding of a Two-Layer Sound--Word Canvas Model for Thai \emph{Klon Pad}}
\author{Worasait Suwannik\thanks{The research reported here was directed by Worasait Suwannik. The programs were produced by Claude, Anthropic's AI model. Worasait Suwannik contributed to the design of the model, including a whole-stanza canvas with a sound layer of pronounced syllables for the form constraints and a word layer of whole words for spelling and naturalness; to the experimental design, including LLM-written topics for reply stanzas; the wording of the cleanness judge prompt; and the unusual-topic test of Appendix~\ref{app:unusual} and the selection of its topics.}}
\date{}

\begin{document}
\maketitle

\begin{abstract}
We present a generator of Thai \emph{klon pad} (กลอนแปด) verse, one stanza from a short topic, and evaluate it against human verse on held-out topics. The generator is a masked language model initialised from WangchanBERTa over a fixed canvas with two layers: a sound layer of pronounced syllables, where rhyme, tone and syllable count are enforced as hard masks, and a word layer of whole words, so that no word is cut into written pieces. Decoding commits one cell at a time and samples at temperature 0.4; one sample is drawn per topic, with no selection, reranking or repair. On 78 held-out topics never used in development, judged by three independent runs of a frozen LLM judge in the same pools as the human stanzas from which the topics were written, 70.5\% of the generated stanzas pass form, lexical cleanness of all four lines and topic alignment together, against 76.9\% for the human stanzas (difference $-6.4$ points, 95\% CI $[-19.2, +6.4]$). Cleanness is level with human (85.9\% vs.\ 87.2\%), and the stanzas pass form and cleanness together slightly more often than the human ones ($+3.8$ points, $[-6.4, +14.1]$); the difference lies in alignment (84.6\% vs.\ 91.0\%), on topics that were written from the human stanzas. The low temperature matters: at higher temperatures the same model passes clearly fewer of the same topics, because lexical errors lie in the model's lower-probability choices; sharpening the sampling distribution removes most of them, at a cost in vocabulary (7.5\% rare words against 16.5\% in the human stanzas) and diversity. The pre-registered target, all-three within 5 points of human, is missed by 1.4 points. KlonSuphap-LM, the only public \emph{klon pad} generator we know of, passes none of the 78 held-out topics when judged in separate pools together with the method's and the human stanzas (form 19.2\%, cleanness 3.8\%).
\end{abstract}

\section{Introduction}

\emph{Klon pad} (กลอนแปด, ``eight-syllable verse'') is the most widely practised form of Thai verse. A stanza (บท, \emph{bot}) has four lines (วรรค, \emph{wak}) of nominally eight syllables, linked by a fixed scheme of external rhymes, with tone constraints on line-final syllables (Section~\ref{sec:form}). Two properties of Thai make it a hard target for a neural generator. Thai script does not mark word boundaries, so a model that works on written subword pieces can produce strings that are not words. And the rhyme is defined on pronounced syllables, which do not always coincide with written ones: ไฉน is read ฉะ|ไหน. A generator therefore has to satisfy position-specific phonological constraints while writing ordinary, well-formed Thai.

This paper describes a generator built for these two properties and evaluates it against human verse on topics it never saw: on 78 held-out topics, 70.5\% of its stanzas pass form, cleanness and topic alignment together, against 76.9\% of the human stanzas from which those topics were written. Its contributions are:
\begin{enumerate}
\item A two-layer canvas model (Section~\ref{sec:method}, Figure~\ref{fig:arch}): one cell per pronounced syllable, on which rhyme, tone and length are hard masks, carrying whole-word tokens over an open vocabulary initialised from WangchanBERTa, decoded by confidence-ordered commitment at a low sampling temperature, one sample per topic.
\item A held-out evaluation (Section~\ref{sec:setting}): 78 topics, written from reply stanzas of test-split threads whose title groups were never used in development, the human stanzas from which the topics were written judged in the same pools, a form checker and two frozen LLM-judge instruments run three times independently.
\item The sampling temperature as the method's main ablation: on the same held-out topics, all-three rises from 44.9\% at temperature 0.8 to 70.5\% at 0.4, cleanness from 56.4\% to 85.9\%, and the number of judged-bad lines falls from 36 to 13 of 312; we explain this by where lexical errors sit in the model's distribution, and measure what it costs in vocabulary and diversity (Sections~\ref{sec:results}--\ref{sec:mechanism}); a confound test attributes the gain to the sampling itself, not to decoder bonuses that temperature also amplifies.
\item A comparison with KlonSuphap-LM \citep{yingseree2023klonsuphap}, run locally with its temperature tuned on development topics, on the same held-out topics, judged in separate pools together with the method's and the human stanzas (Section~\ref{sec:baseline}).
\end{enumerate}

The trained model is available on the Hugging Face Hub at \url{https://huggingface.co/voraseth/klon8-base}, together with the code that generates from it, and loads with the Transformers library \citep{wolf2020transformers}:
\begin{verbatim}
from transformers import AutoModel
model = AutoModel.from_pretrained("voraseth/klon8-base", trust_remote_code=True)
print(model.generate_poem("topic in Thai")[0]["poem"])
\end{verbatim}

\section{Related Work}
\label{sec:related}

\paragraph{Thai poetry generation.} Published work on generating Thai verse is sparse. The only public \emph{klon pad} generator we know of is KlonSuphap-LM \citep{yingseree2023klonsuphap}, a Thai GPT-2 base model fine-tuned on about 65,770 couplets (บาท, two lines each) of \emph{klon pad} from the Thai Literature Corpora \citep{sawatphol2019tlc}. It marks the words that must rhyme with special tags in the training text (a tag pair for the source and the target of each rhyme within a stanza, and another for the rhyme between stanzas), optionally masks attention to non-rhyme words, and then fine-tunes with proximal policy optimisation (PPO) against a rule-based form grader. It is documented by a model card and code repository rather than a paper, and reports no evaluation of meaning or fluency; form is learned softly from the tags and the reward, whereas our decoder enforces rhyme, tone and length as hard masks on pronounced syllables. The ThaiProsody benchmark \citep{yongsatianchot2026composing} evaluates large language models composing four Thai verse forms, \emph{klon pad} among them, with a deterministic rule checker for tone, weight and rhyme, a model-based syllable counter and an LLM content judge; it is a benchmark of general-purpose models prompted zero-shot rather than a trained generator. Public Thai poem collections include the Thai Literature Corpora \citep{sawatphol2019tlc}, drawn from classical works, and the klongklon dataset \citep{pythainlp2023klongklon}, more than 50,000 \emph{klon pad} and \emph{khlong si suphap} poems collected from various literature, among them classical works such as Phra Aphai Mani; our training data are instead modern verse written by forum members, and our evaluation compares against the human stanzas from which its topics were written. We are not aware of a published \emph{klon pad} generator evaluated against human verse on held-out topics; we evaluate KlonSuphap-LM under our protocol in Section~\ref{sec:baseline}.

\paragraph{Thai language models.} The generator's body is WangchanBERTa \citep{lowphansirikul2021wangchanberta}, a RoBERTa-style masked language model pretrained on Thai text with a SentencePiece vocabulary. Segmentation and grapheme-to-phoneme conversion use PyThaiNLP \citep{phatthiyaphaibun2023pythainlp} and TLTK \citep{tltk}. We use a masked encoder rather than a left-to-right generative model because the canvas is filled in any order under position-specific constraints, and because WangchanBERTa's pieces give an initialisation for every whole word of the open vocabulary.

\paragraph{Constrained poetry and lyric generation.} Neural generation of metrical verse has concentrated on Chinese and English. Hafez \citep{ghazvininejad2016hafez,ghazvininejad2017hafez} combined a language model with a finite-state acceptor that admits only metrical, rhyming paths, the closest antecedent of our hard masks. \citet{tian2022sonnet} plan content and rhyme words before writing sonnet lines. For rigid-format text, SongNet \citep{li2020songnet}, motivated by lyrics and evaluated on SongCi and sonnets, controls the format with format and rhyme symbols at each position, a fixed grid like our canvas; for lyrics, DeepRapper \citep{xue2021deeprapper} writes rap lines right to left so that the rhyme is chosen first. CIGEN \citep{song2022cigen}, the work nearest in spirit to ours, composes Chinese Ci non-autoregressively so that generation accommodates the tune pattern's rigid format of length and rhyme. PoetryDiffusion \citep{hu2024poetrydiffusion} writes sonnets and SongCi with a continuous diffusion model, which noises continuous word representations towards a Gaussian, for the semantics, and a separately trained metrical controller that penalises states straying from the target control distribution; its form control is soft, where ours is a hard mask. A shared design question is when the rhyme word is chosen relative to the rest of the line. Our decoder chooses the rime family early from the model's own probability over rime families (Section~\ref{sec:decoding}), writes it into the sound layer, and lets the words of the whole stanza be filled around it in confidence order. A second difference is the unit: because Thai rhyme lives on pronounced syllables and Thai script does not delimit words, we keep two aligned layers, sounds for the constraints and whole words for the text, instead of constraining written subword tokens. Sound has also been given to masked language models as extra input embeddings: PoemBERT \citep{huang2025poembert} adds sentiment and pinyin embeddings to a BERT-based Chinese poem generator. Our sound channels are also input embeddings, but the sound layer is in addition the surface on which the form constraints act.

\paragraph{Constrained and non-autoregressive decoding.} Lexically constrained beam search \citep{hokamp2017grid} and NeuroLogic decoding \citep{lu2021neurologic} enforce constraints in left-to-right generation. Our generator is a bidirectional masked language model \citep{devlin2019bert} used non-autoregressively. Mask-Predict \citep{ghazvininejad2019maskpredict} and MaskGIT \citep{chang2022maskgit} fill a fixed-length canvas over several steps. MaskGIT keeps the most confident predictions at each step and re-predicts the rest; Mask-Predict predicts all tokens at every step and re-masks the least confident ones, so earlier predictions can change. Our decoder instead commits one cell per step, the open cell whose best admissible token is most confident, and does not re-mask committed cells. Our training, which masks a randomly drawn share of each stanza (Section~\ref{sec:method}), is close to absorbing-state discrete diffusion: in D3PM an absorbing ([MASK]) state reduces to a reweighted masked-LM objective \citep{austin2021d3pm}, and the MDLM objective is a weighted average of masked-LM losses \citep{sahoo2024mdlm}. These models do not order unmasking by the model's confidence at decoding time; that ordering is the one of Mask-Predict and MaskGIT. The difference is that our constraints are hard masks on a parallel sound layer, so every intermediate canvas stays inside the form.

\paragraph{Sampling temperature.} \citet{holtzman2020degeneration} showed that the low-probability tail of a language model is unreliable and that pure sampling from it produces incoherent text, motivating nucleus sampling. \citet{peeperkorn2024temperature} found that in story generation temperature is only weakly related to novelty and moderately related to incoherence. Our temperature comparison is a measurement of the same trade-off in a constrained, non-autoregressive generator, against human verse on held-out topics: the tail holds most lexical errors, and sharpening the distribution removes them at a measurable cost in vocabulary. Best-of-$N$ selection with a learned verifier \citep{cobbe2021verifiers} is the usual alternative for avoiding bad samples after the fact; the method here draws one sample with no selection.

\paragraph{LLM judges.} LLMs have been used as judges of generated text since early studies \citep{chiang2023llmeval,zheng2023judging}, with known position and verbosity biases \citep{zheng2023judging}; for creative writing, LLM judgments agree poorly with expert readers \citep{chakrabarty2024art}. Our cleanness instrument adapts the error-span protocol of GEMBA-MQM \citep{kocmi2023gemba} and asks a narrow question (is each span well-formed Thai) rather than a judgment of quality; we add fixed human anchor stanzas in every pool, blind interleaving with human verse, three independent judge runs, and report the judge's disagreement with a human reader as a limitation (Section~\ref{sec:limits}).

\section{\emph{Klon Pad}, Data and Evaluation Setting}
\label{sec:setting}

\subsection{The skeleton of a \emph{klon pad} stanza}
\label{sec:form}

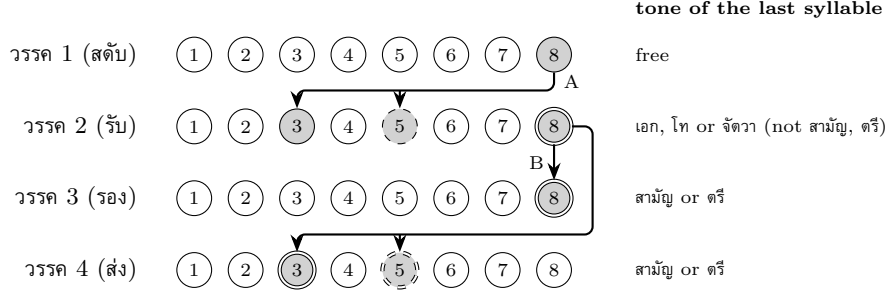
\begin{figure}[t]
\centering
\begin{tikzpicture}[
  font=\small, >=Stealth,
  syl/.style={circle, draw, minimum size=0.46cm, inner sep=0pt, font=\scriptsize},
  ga/.style={syl, fill=gray!35},
  gb/.style={syl, fill=gray!35, double, double distance=0.8pt},
  alt/.style={dashed},
  link/.style={->, thick, rounded corners=3pt}]
\def\X#1{#1*0.68+0.4}
\foreach \r/\role in {1/สดับ, 2/รับ, 3/รอง, 4/ส่ง}
  \node[anchor=east] at (0.45, -\r*0.95) {วรรค~\r\ (\role)};
\foreach \c in {1,...,7} \node[syl] (a\c) at ({\X{\c}}, -0.95) {\c};
\node[ga] (a8) at ({\X{8}}, -0.95) {8};
\foreach \c in {1,2,4,6,7} \node[syl] (b\c) at ({\X{\c}}, -1.90) {\c};
\node[ga] (b3) at ({\X{3}}, -1.90) {3};
\node[ga, alt] (b5) at ({\X{5}}, -1.90) {5};
\node[gb] (b8) at ({\X{8}}, -1.90) {8};
\foreach \c in {1,...,7} \node[syl] (c\c) at ({\X{\c}}, -2.85) {\c};
\node[gb] (c8) at ({\X{8}}, -2.85) {8};
\foreach \c in {1,2,4,6,7,8} \node[syl] (d\c) at ({\X{\c}}, -3.80) {\c};
\node[gb] (d3) at ({\X{3}}, -3.80) {3};
\node[gb, alt] (d5) at ({\X{5}}, -3.80) {5};
\draw[link] (a8.south) |- node[pos=0.25, right, font=\scriptsize]{A} ({\X{3}}, -1.425) -- (b3.north);
\draw[link, alt] ({\X{5}}, -1.425) -- (b5.north);
\draw[link] (b8.south) -- node[left, font=\scriptsize]{B} (c8.north);
\draw[link] (b8.east) -- ({\X{8.75}}, -1.90) |- ({\X{3}}, -3.33) -- (d3.north);
\draw[link, alt] ({\X{5}}, -3.33) -- (d5.north);
\node[anchor=west, font=\scriptsize\bfseries] at ({\X{9.4}}, -0.35) {tone of the last syllable};
\node[anchor=west, font=\scriptsize] at ({\X{9.4}}, -0.95) {free};
\node[anchor=west, font=\scriptsize] at ({\X{9.4}}, -1.90) {เอก, โท or จัตวา (not สามัญ, ตรี)};
\node[anchor=west, font=\scriptsize] at ({\X{9.4}}, -2.85) {สามัญ or ตรี};
\node[anchor=west, font=\scriptsize] at ({\X{9.4}}, -3.80) {สามัญ or ตรี};
\end{tikzpicture}
\caption{Skeleton of one \emph{klon pad} stanza as checked in this work. Circles are pronounced syllables, drawn at the usual eight per line. Group A links the last syllable of line~1 to the 3rd syllable of line~2 (or the 5th, dashed); group B links the last syllable of line~2 to the last of line~3 and to the 3rd (or 5th) syllable of line~4. Line-final tones are on the right.}
\label{fig:skeleton}
\end{figure}

A stanza (บท) has four lines (วรรค), traditionally named วรรคสดับ, วรรครับ, วรรครอง and วรรคส่ง. Each line has 7--9 pronounced syllables, usually 8; in the training stanzas 73--81\% of lines have 8 and 18--24\% have 9, depending on the line. The form checker used throughout this work (distributed with the model) reads a stanza as follows; Figure~\ref{fig:skeleton} shows the skeleton.
\begin{itemize}
\item \textbf{Syllables.} Each line is segmented into words with PyThaiNLP \citep{pythainlp} using its default \texttt{newmm} tokenizer, a dictionary-based maximum-matching segmenter \citep{phatthiyaphaibun2023pythainlp}, and each word is converted to phonemes with TLTK \citep{tltk}, giving pronounced syllables with onset, vowel, coda and tone. A line passes the count when it has 7--9 syllables, where short unstressed syllables (\emph{e.g.}\ สะ, กระ) may count as one, as half or not at all; a line of 10 pronounced syllables that contains such light syllables can therefore pass, which is why the training data and the canvas below allow up to 10 syllables per line.
\item \textbf{External rhyme.} Two syllables rhyme when they have the same \emph{rime}, the vowel and the final consonant (rimes are grouped into \emph{rime families}, with the ไ/ใ/-ัย \textasciitilde\ -าย and เ-า \textasciitilde\ -าว families merged, because the phoneme converter is inconsistent about their length). Three links are required: the last syllable of line~1 with the 3rd or 5th syllable of line~2 (group A); the last syllable of line~2 with the last of line~3; and the last syllable of line~2 with the 3rd or 5th syllable of line~4 (group B). Since lines 2 and 3 end on the same rime, the last link is equivalently a rhyme between the end of line~3 and the 3rd or 5th syllable of line~4. The link from the end of line~4 to the end of line~2 of the next stanza applies only to multi-stanza poems and is not used here.
\item \textbf{Repeated rhyme word.} A rhyme link may not be made by repeating the same syllable of the same word.
\item \textbf{Tones.} The last syllable of line~2 must not carry the สามัญ (mid) or ตรี (high) tone, so it carries เอก, โท or จัตวา; the last syllables of lines 3 and 4 must carry สามัญ or ตรี. The end of line~1 is free.
\end{itemize}
A stanza passes form when all four counts, the three links and the three tone rules hold, no rhyme word is repeated, and it contains no Chinese, Japanese or Korean script (a guard for generators with multilingual vocabularies; it never fires on the method, whose vocabulary is Thai only). Internal rhymes (สัมผัสใน) are an ornament and are not required. Poets bend these rules: 92.3\% of the human stanzas in the held-out evaluation pass the checker, and 95.0\% of the training stanzas.

\subsection{Training data}
\label{sec:data}

The corpus is modern verse written by the users of a Thai poetry forum and collected from its public boards, klonthaiclub.com (บ้านกลอนไทย), whose 29 boards used here cover love, longing, encouragement, festivals, dharma and a practice room for the prosody rules. Posts are cut into four-line stanzas; a stanza is kept when its line count and syllable counts are well formed and all three external rhyme links hold (tones are not filtered), and duplicates are removed. The training split has 50,788 stanzas from 3,956 threads: 11,601 from opening posts and 39,187 from replies. The data pipeline of the two-layer model further keeps a stanza only when every line has 7--10 pronounced syllables, every syllable is read by the phoneme converter, and the words, re-joined and re-segmented, give back exactly the same sounds; 49,194 stanzas remain, of which a random 1,000 are held out to monitor the validation loss and 48,194 are used for training.

The conditioning topic of every training stanza, opening post or reply, is its thread title after removing decoration (symbol runs, repeated punctuation, reply prefixes, bracketed notes, decorative Latin letters).

\subsection{Train/test split and the held-out topics}
\label{sec:split}

The corpus is split by thread title, not by stanza. Titles are first grouped on their Thai-only normalisation (the \emph{title key}), so that decorated duplicates of one title (ฝันร้าย and the same title framed by star symbols) cannot fall on both sides. The split has 3,821 training titles in 3,700 title groups (50,788 stanzas) and 659 test titles in 652 title groups (8,680 stanzas). A stanza-level split would leak, because a title's content syllables occur in a sibling stanza of the same thread 22.4\% of the time, as often as in the stanza itself (22.5\%).

For the evaluation, every test title group that had been used anywhere in development was excluded. A group was out if its title key matched any title, prompt or topic used anywhere in development (every prompt file, every judged pool, every judged batch), or if any of its threads had supplied a human stanza to a judged pool or had a line found in a judged batch. The exclusion list holds 4,856 title keys from all these sources and 396 test threads (388 threads of human stanzas used in earlier pools, together with the threads found by their lines). Of the 652 test title groups (690 threads), 552 were excluded, leaving 100 (110 threads, 1,279 stanzas). None of the 100 titles was usable as a prompt as it stood; the reasons included birthday greetings, contest announcements, threads of ornamental \emph{klon bot} forms, titles with Latin letters and one-syllable titles. A fixed rule therefore took, for each of the 100 title groups, its first reply stanza whose topic, written from that stanza by Claude Opus 5.5 under a fixed guide, is 2--12 syllables of Thai only (no Latin letters, digits or stray combining marks), contains none of a fixed list of forum keywords (meta and event words such as วันเกิด, ประกาศ, สมาชิก, กระทู้) and is neither a title or prompt used in development, a training title, nor a topic that Claude wrote for training stanzas during development (the model evaluated here was trained on thread titles only). That stanza is the \textbf{human control} and its topic is the \textbf{prompt} for the generator as well; below, the \emph{held-out topics} are these prompts, one per title group (Section~\ref{sec:limits} discusses how this selection may bias alignment). 78 title groups qualified and 22 were dropped. Development used 49 further test-split titles, the \emph{49 development topics} below, which are therefore among the excluded groups. When the evaluation ran, the exclusion of the 78 held-out topics' title groups and their human controls was checked again against a fresh scan of every prompt, pool and judged line (the 388 pool threads and 53,728 judged lines): zero overlap.

Because each prompt was written from its human control, it may fit that stanza better than any other poem, which favours the human arm on alignment. Two further human stanzas per topic were judged in the same pools as fillers; they balance the pools and are never an endpoint. Where the title had enough replies (138 of the 156 filler slots) they are the two earliest other reply stanzas under the same title (6 of them from a second thread with the same title), each under its own Claude-written topic; the other 18 slots hold human controls of the second earlier evaluation (Section~\ref{sec:criteria}), 11 reply stanzas under Claude-written topics and 7 opening-post stanzas under their thread titles. Fillers are exempt from the exclusion rule because they are never endpoints. The human controls are verse written by forum users: 17 of the 78 come from the forum's birthday-verse board.

\subsection{Passing criteria and the judge}
\label{sec:criteria}
\label{sec:judge}

A stanza passes \textbf{all-three} when (i) it passes the form checker of Section~\ref{sec:form}, (ii) all four of its lines are judged clean, and (iii) its topic alignment is judged at least 4 of 5. The marginal rates are reported beside the conjunction.

\paragraph{Target.} Our target was fixed in a pre-registration before the evaluation reported here: on held-out topics, the all-three rate of the generated stanzas should be within 5 percentage points of the rate for the human stanzas of the same topics, judged in the same pools. Two earlier held-out evaluations with the same passing criteria, judge instruments and target, of predecessor configurations that drew 64 samples per prompt at temperature 0.8 and kept the one an LLM ranker preferred (the first with an earlier model and one further step that re-drew the weakest line, the second with the model evaluated here), each on its own fresh title groups and human controls (thread titles with opening-post stanzas in the first; in the second, 41 such titles and 40 Claude-written topics of reply stanzas) and with three independent judge runs, missed it widely (44.7\% against 63.1\%, and 35.8\% against 60.5\%), in both cases because of cleanness; the human rate differs between evaluations because each has its own prompts, human stanzas and pools. The system described here comes within 6.4 points (70.5\% against 76.9\%, Section~\ref{sec:results}): the target is missed by 1.4 points, and the 95\% interval of the difference, $[-19.2, +6.4]$, is too wide to separate the system from human verse in either direction.

\paragraph{Cleanness.} The cleanness instrument (Appendix~\ref{app:prompts}) adapts the error-span protocol of GEMBA-MQM \citep{kocmi2023gemba}. The judge marks exact spans that are not linguistically well formed Thai: non-words, unrecognisable misspellings, unnatural word formations, and unnatural combinations of words that disrupt reading. A span is \emph{critical} if it makes the line incomprehensible and \emph{major} if a native reader stumbles but understands, or if it is a non-word other than a genuine Thai prefix or suffix that can attach to the neighbouring word; something technically odd that does not disrupt reading is a pass. Because the text is verse, abstraction, metaphor, personification, archaic or poetic diction, unusual imagery and natural poetic compounds are explicitly not errors, and rhyme, syllable count and structure are not judged. A line is clean when it carries no critical or major span.

\paragraph{Alignment.} The alignment instrument (Appendix~\ref{app:prompts}) asks the judge to write one Thai sentence saying what the poem is about, then score 1--5 how much of it stays on the topic, judging subject matter only and accepting figurative treatment (5: every line on topic; 4: at most one line wanders; 3: partly; 2: touched once and left; 1: unrelated). A stanza is aligned when its score is at least 4.

\paragraph{Runs, aggregation and agreement.} The judge is Claude Opus 5.5, run as an agent: a conversation whose only tool use is reading the batch files and writing its verdict file. Both prompts are frozen and wrapped in a fixed agent instruction (Appendix~\ref{app:prompts}). Each instrument is run three times independently, each run judging all six pools of the instrument in one conversation (Section~\ref{sec:limits}): a line is clean when at least two of the three runs report no critical or major span on it, and alignment is the median of the three scores; an item missing from a run counts in that run as unclean on every line and as alignment 1. Fleiss' kappa across the three runs, over all items except the anchors, is 0.76 for per-line cleanness and 0.84 for alignment $\geq 4$.

\paragraph{Pools, anchors and blinding.} The 78 topics are split into six parts of 13. Each part has one cleanness pool and one alignment pool holding, for each topic, the three system arms (temperatures 0.4, 0.6 and 0.8), the human control and the two fillers, plus 16 \emph{anchor} stanzas: human stanzas from a fixed anchor set, the same in every pool, used to detect judge drift (about 17\% of items). Items are interleaved and keyed at random and dealt into seven batch files per pool of 11 or 12 items plus two or three anchors, texts are given raw, and the agent instruction tells the judge that the pool mixes system and human poems without labels. A pool-run is void if its anchors are under 70\% clean or their median alignment is under 3; none was. Anchors were clean in 100\%, 87.5\% and 81.3\% of cases in runs 1, 2 and 3.

\paragraph{Statistics.} The primary comparison, fixed before any judging, is all-three at temperature 0.4 against the human controls; the other temperatures, the marginals, kappa and vocabulary and diversity read-outs are secondary. Differences use topic-clustered percentile bootstrap intervals (10,000 resamples of topics, seed 0, arms resampled together), and discordant topics are tested with the exact McNemar test. The McNemar tests and discordant-topic counts, the ratios of rates, the per-arm intervals, the comparison of temperatures 0.6 and 0.4, the bad-line counts of Table~\ref{tab:bad}, the originality check of Section~\ref{sec:originality} and the rime-family count of Section~\ref{sec:mechanism} were added after judging and are not part of the pre-registration.

\section{Method}
\label{sec:method}

\begin{figure}[t]
\centering
\begin{tikzpicture}[
  font=\footnotesize, >=Stealth, node distance=0.3cm,
  blk/.style={draw, rounded corners=2pt, align=left, text width=7.0cm, inner sep=3pt, fill=white},
  dec/.style={draw, rounded corners=2pt, align=left, text width=7.0cm, inner sep=3pt, fill=gray!8},
  cellrow/.style={draw, minimum height=0.5cm, inner sep=1pt, font=\scriptsize}]
\node[cellrow, minimum width=0.6cm, anchor=west] (bos) at (0,0) {\texttt{<s>}};
\node[cellrow, minimum width=1.6cm, right=0pt of bos, fill=gray!20] (tp) {topic: 12 cells};
\node[cellrow, minimum width=1.25cm, right=0pt of tp] (w1) {วรรค 1: 10};
\node[cellrow, minimum width=1.25cm, right=0pt of w1] (w2) {วรรค 2: 10};
\node[cellrow, minimum width=1.25cm, right=0pt of w2] (w3) {วรรค 3: 10};
\node[cellrow, minimum width=1.25cm, right=0pt of w3] (w4) {วรรค 4: 10};
\node[anchor=north west, align=left, text width=7.0cm, font=\scriptsize, inner sep=1pt] at ([yshift=-2pt]bos.south west)
  {\textbf{Canvas}: one cell per pronounced syllable; topic cells hold the topic's words, poem cells start as \texttt{<mask>}.};
\node[blk, anchor=south west] (emb) at ([yshift=0.45cm]bos.north west) {%
  \textbf{Cell input} $=$ word layer $+$ sound layer $+$ position.\\
  \emph{Word layer}: token $(w,k)$, the $k$-th syllable of word $w$ with $n_w$ syllables, embedded as $e_w + e_{(k,n_w)}$; 62{,}665 words, each row initialised from the mean WangchanBERTa embedding of the word's pieces (37{,}851 words never seen in the poems: rows frozen, one shared bias offset).\\
  \emph{Sound layer}: onset $+$ rime $+$ tone embeddings of the cell's syllable (filled cells: from the token; a masked rhyme cell: the chosen rime).\\
  \emph{Position}: line, cell and role embeddings $+$ WangchanBERTa positions.};
\node[blk, above=of emb] (body) {\textbf{WangchanBERTa-base encoder}: pretrained, 12 layers, width 768; a leading \texttt{<s>} cell, padding never attended.};
\node[blk, above=of body] (head) {\textbf{Pretrained masked-LM head transform}: $z_i$ (dense, GELU, LayerNorm).};
\node[blk, above=of head] (out) {\textbf{Word-slot logits} at cell $i$: $z_i\cdot e_w + z_i\cdot e_{(k,n_w)} + b_w$, tied to the input rows. Training: masked prediction.};
\foreach \n in {bos,tp,w1,w2,w3,w4} \draw[->] (\n.north) -- (\n.north |- emb.south);
\draw[->] (emb) -- (body); \draw[->] (body) -- (head); \draw[->] (head) -- (out);
\node[dec, anchor=north west] (m) at ([xshift=0.6cm]out.north east) {%
  \textbf{1. Scores}: $+1.5$ on the topic's words (cleared for a word in a line once it is used there), $-2$ on words already placed; hard masks: a line's first / last cell takes only a word's first / last syllable (word layer); line-final tones (Figure~\ref{fig:skeleton}) and the stutter ban (sound layer).};
\node[dec, below=of m] (s) {%
  \textbf{2. At step 5 only}: each rhyme group's rime family is drawn from the mass that its line-final and 3rd-syllable cells give to each family (temperature 0.5; the groups get different families); the receiver is the 3rd or 5th syllable, whichever has more mass; the rime is written into the group's open cells, which take only that rime from then on (no word twice in a group). Until then rhyme cells are held.};
\node[dec, below=of s] (c) {%
  \textbf{3. Top 512}: only the 512 best tokens per cell are kept.};
\node[dec, below=of c] (f) {%
  \textbf{4. Commit one cell}: the open cell with the most confident admissible token; sample $(w,k) \propto \exp(\text{log-prob at the cell}/T)$ with $T = 0.4$; the rest of $w$'s span is forced, so every syllable is part of a word.};
\draw[->] (out.east) -- (m.west |- out.east);
\draw[->] (m) -- (s); \draw[->] (s) -- (c); \draw[->] (c) -- (f);
\draw[->, thick] (f.south) |- (w4.east);
\node[anchor=north west, align=left, text width=3.2cm, font=\scriptsize] at ([xshift=4pt, yshift=-2pt]f.south) {write the token and its sound into the canvas; repeat until every poem cell is filled};
\end{tikzpicture}
\caption{The generator. Left: the network, a WangchanBERTa encoder over a canvas of topic cells and $4 \times 10$ poem cells, each cell carrying a whole-word token (word layer) and its pronounced syllable (sound layer). Right: one decoding step, repeated until every poem cell is filled (one step per cell, 32.9 on average on the held-out topics); rhyme, tone and length masks act on the sound layer; word-boundary masks and the sampling act on the word layer.}
\label{fig:arch}
\end{figure}
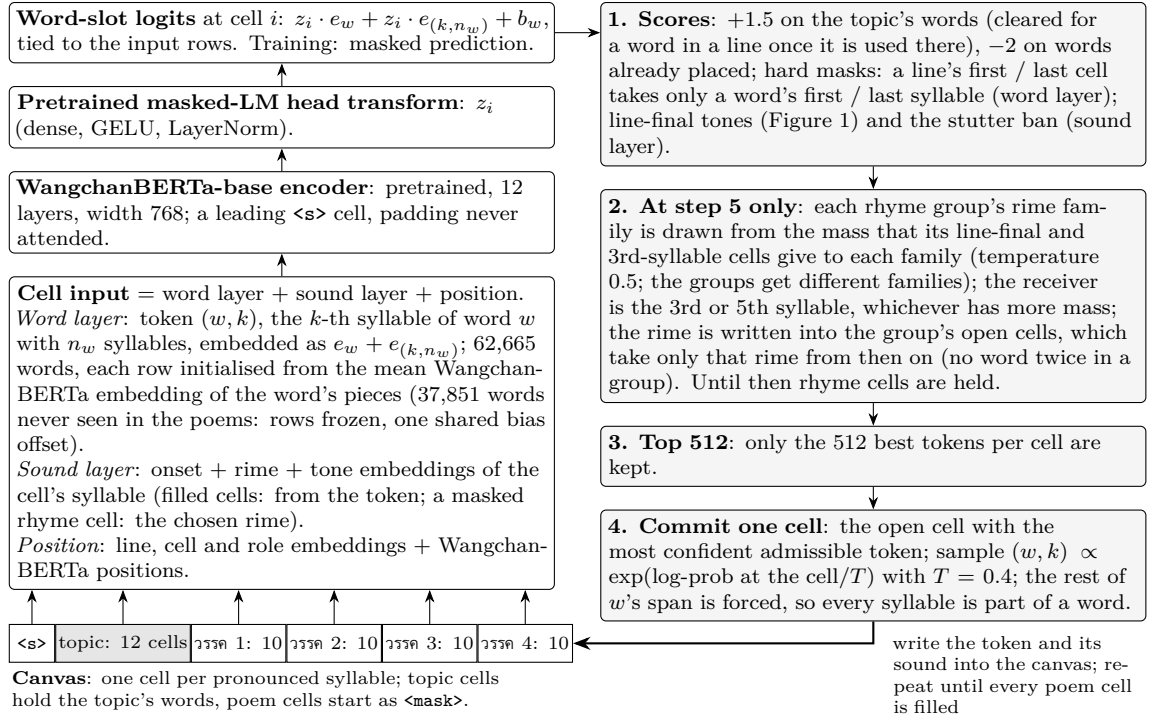

\subsection{Two-layer canvas model}

A stanza is a fixed canvas of cells preceded by the topic: a leading \texttt{<s>} cell, 12 topic cells and four lines of 10 cells (Figure~\ref{fig:arch}). The model has two aligned layers.

\paragraph{Sound layer.} One cell per pronounced syllable, computed as the form checker computes it (\texttt{newmm} words, then TLTK grapheme-to-phoneme per word), carrying three channels: onset, rime family and tone. The syllable count of a line is its number of cells, and rhyme and tone masks act on these sounds, so ไฉน fills two cells, ฉะ and ไหน, and its rhyme is that of ไหน. In the model used here the sound of a filled cell is derived from its word-layer token; a masked cell has no sound input, except that a rhyme cell whose rime family has been chosen receives that rime in its rime channel.

\paragraph{Word layer.} The token at a cell is a pair $(w,k)$: word $w$ and the position $k$ of the cell's syllable within it, so a word is never cut into written pieces. Its input embedding is $e_w + e_{(k,n_w)}$, where $n_w$ is the number of syllables of $w$, plus the embeddings of the three sound channels, of the line (topic or poem line 1--4), of the cell's position within its line and of its role (topic cell or poem cell of a given line), plus WangchanBERTa's pretrained position embeddings; the sum passes through WangchanBERTa's embedding layer normalisation. The vocabulary is open: 62,665 words, 24,814 seen in the 49,194 stanzas of the data pipeline (training and validation) and 37,851 never seen in them (from PyThaiNLP's word list and whole-word WangchanBERTa pieces; each of these is kept only if it is Thai, readable by the phoneme converter, at most 8 syllables and re-segmented as itself). Every word row starts from the mean WangchanBERTa embedding of its pieces (86\% of word tokens are exactly one piece); rows of never-seen words stay frozen, with one learned bias offset for the group. On test stanzas the open vocabulary lowers the share of stanzas with an out-of-vocabulary word from 14.0\% (with the training-stanza words alone) to 2.3\%.

\paragraph{Body and output.} The body is the pretrained WangchanBERTa-base encoder \citep{lowphansirikul2021wangchanberta} (12 layers, width 768); padding is never attended, and the \texttt{<s>} cell is always attended. Its hidden state at a cell passes through WangchanBERTa's pretrained masked-LM head transform to $z_i$, and the logit of token $(w,k)$ is $z_i\cdot e_w + z_i\cdot e_{(k,n_w)} + b_w$, tied to the input rows, where $b_w$ is a per-word output bias initialised from the centred, add-one-smoothed log frequency of $w$ in those stanzas (frozen for never-seen words, which share the learned offset above). The \emph{word-slot tokens} are all pairs $(w,k)$ with $k \le n_w$, 164,818 in all (164,822 output tokens with four special tokens); the cross-entropy over all output tokens is computed exactly, without forming every logit, by factorising its normaliser over words.

\paragraph{Training.} Training is masked prediction with one masking scheme per stanza: whole words (40\%, a share of the words drawn uniformly from $[0.15, 1]$), single cells (20\%, a share of the cells drawn the same way), lines $j$ to 4 for a start line $j$ drawn uniformly from 1--4 (25\%) and every cell (15\%). At masked rhyme cells the rime is revealed in the sound channel with probability 0.5, matching decoding, where the rime family is chosen before the word. The model was trained for 24 epochs with AdamW (weight decay 0.01 on all parameters except the per-word output biases, the shared offset, the \texttt{<s>} cell vector and the pretrained position embeddings), batch 32 and a one-cycle learning-rate schedule peaking at $3\times10^{-5}$ after 6\% of the 36,144 steps, with gradient clipping at 1.0 and seed 0; the weights after the last epoch are used. Training took about 8.7 hours on an Apple-silicon GPU. In development, 24 rather than 12 epochs raised all-three by 8.2 points (95\% CI $[+2.9, +13.7]$; one judge run, single samples with no selection on the 49 development topics with ten seeds each).

\subsection{Decoding}
\label{sec:decoding}

Decoding starts from a canvas whose poem cells are masked, with a length for each line sampled from a table of line lengths measured on human stanzas (7--10 syllables, mostly 8). Each step runs the model once and then (Figure~\ref{fig:arch}, right):
\begin{enumerate}
\item scores every open cell: to the model's logits it adds a bonus of $+1.5$ on the tokens of the topic's words (the lexicon words of more than one character among the 12 topic cells), cleared for a topic word in a line once that word is completed there, and a penalty of $-2$ on words already placed; it applies the hard masks: the first cell of a line admits only the first syllable of a word and the last cell only the last syllable, line-final cells admit only the allowed tones (Figure~\ref{fig:skeleton}), and a stutter ban forbids a token whose sound equals its neighbour's in the same line unless the two cells belong to one lexicon word or the word pair is attested as a same-sound join at least twice in the 49,194 stanzas of the data pipeline (602 of 1,613 pair types are allowed, e.g.\ รัก|รัก, นาน|นาน); each cell is normalised to log-probabilities before the stutter ban (and, once chosen, the rime-family and group-word masks of step~2) remove tokens, without renormalising;
\item at the fifth step only, once four cells have been committed, chooses each rhyme group's rime family: family $f$ is drawn with probability proportional to $\exp(s_f/0.5)$, where $s_f$ is the mean, over the group's line-final cells and its 3rd-syllable receiver, of the log of the probability mass that the cell gives to $f$ as scored in step~1; the family temperature 0.5 is fixed and $T$ does not affect it, and the two groups always receive different families. The receiving cell is the 5th syllable if the chosen rime has more mass there than at the 3rd, and otherwise the 3rd; when it is the 5th, the 3rd may not carry that rime. The rime is written into the rime channel of the group's open cells, so that the words are then chosen to fit it, and from then on rhyme cells admit only their group's rime and no word may appear twice in a rhyme group. Until this step the rhyme cells are held, so the first four steps write only other cells;
\item keeps only the 512 best tokens of each cell;
\item commits one cell: the open cell whose best admissible token has the highest log-probability. A token $(w,k)$ is admissible only if the whole span of $w$ fits in the line around that cell and every other cell of the span admits the matching syllable of $w$. The token is sampled with probability proportional to $\exp(s/T)$, where $s$ is its log-probability at that cell and $T = 0.4$; the other cells of $w$'s span are then forced to the matching syllables of $w$ and committed in later steps, so every syllable belongs to a real word.
\end{enumerate}
A stanza takes one forward pass per syllable cell, 32.9 on average on the held-out topics (31--36). If the checker's re-segmentation of the joined words disagrees with the cells, or no open cell has an admissible token, the stanza is decoded again, continuing the same random stream; this validity retry is allowed at most four times, after which the last attempt is kept, and at $T=0.4$ it never occurred on the 78 held-out topics. Apart from it, one sample is drawn per topic (seed 0); there is no best-of-$N$, no reranker, no rejection on quality and no line repair. On one Apple-silicon laptop GPU the 78 held-out stanzas took 126 seconds including model loading, about 1.6 seconds per stanza.

\section{Results}
\label{sec:results}

\begin{table}[t]
\centering
\caption{Held-out evaluation: 78 topics, one sample per topic and temperature, three independent judge runs per instrument, all arms and the human control in the same pools. Differences are against human with 95\% topic-clustered bootstrap intervals; pp = percentage points. The fillers row gives the first and the second filler of each topic.}
\label{tab:confirm}
\small\setlength{\tabcolsep}{4pt}
\resizebox{\linewidth}{!}{%
\begin{tabular}{lrrrrrr}
\toprule
Arm & All three & vs.\ human (pp) & Clean & Aligned $\geq 4$ & Form & Clean \& form \\
\midrule
$T=0.4$ (the method) & 70.5\% & $-6.4$ [$-19.2$, $+6.4$] & 85.9\% & 84.6\% & 98.7\% & 84.6\% \\
$T=0.6$ & 55.1\% & $-21.8$ [$-35.9$, $-7.7$] & 65.4\% & 79.5\% & 98.7\% & 64.1\% \\
$T=0.8$ & 44.9\% & $-32.1$ [$-46.2$, $-16.7$] & 56.4\% & 74.4\% & 100.0\% & 56.4\% \\
Human, same topics & 76.9\% & -- & 87.2\% & 91.0\% & 92.3\% & 80.8\% \\
\midrule
Fillers 1 / 2 (human, not endpoints) & 83.3 / 79.5\% & -- & 88.5 / 91.0\% & 98.7 / 93.6\% & 94.9 / 94.9\% & 83.3 / 85.9\% \\
\bottomrule
\end{tabular}}
\end{table}

\paragraph{Against human verse.} The method passes all-three on 55 of 78 topics (70.5\%, 95\% CI $[60.3, 80.8]$) and the human stanzas on 60 (76.9\%, $[67.9, 85.9]$). The paired difference is $-6.4$ points ($[-19.2, +6.4]$); the method passes where the human stanza fails on 11 topics and the reverse holds on 16 (exact McNemar $p = 0.44$). As a share of the human rate the method reaches 0.917 ($[0.762, 1.091]$). Table~\ref{tab:confirm} gives all rates; Appendix~\ref{app:poems} shows 30 of the 55 passing stanzas.

\paragraph{Where the difference lies.} The method is 6.4 points below human on alignment ($[-17.9, +3.8]$; 7 vs.\ 12 discordant topics) and 3.8 points above on cleanness and form together ($[-6.4, +14.1]$; 11 vs.\ 8). On cleanness alone it is 1.3 points below ($[-11.5, +9.0]$; 7 vs.\ 8). It passes form more often than the human stanzas ($+6.4$, $[0.0, +12.8]$), because poets bend the checker's rules. The alignment difference is in the direction predicted by the bias from writing each prompt from its human control (Section~\ref{sec:split}); the fillers, also judged against topics written from themselves (or, for 7 opening-post stanzas, against the thread title they were written to), reach 98.7\% and 93.6\% alignment. We read this as evidence that the method is level with human verse in this pool on cleanness, and below it on alignment by an amount that the topic-writing procedure may partly or wholly explain.

\paragraph{Ratios.} Relative to the all-three rate at $T=0.4$, the same model and decoder reach 0.782 of it at temperature 0.6 (95\% CI $[0.652, 0.923]$) and 0.636 at 0.8 ($[0.489, 0.804]$).

\paragraph{Temperature, on the same topics.} The same model and decoder at temperature 0.8 pass 44.9\%, and at 0.6 55.1\%. The step from 0.8 to 0.4 is $+25.6$ points of all-three ($[+12.8, +38.5]$; 25 vs.\ 5 topics, $p = 0.0003$) and $+29.5$ of cleanness ($[+17.9, +41.0]$; 26 vs.\ 3), with alignment $+10.3$ ($[0.0, +20.5]$; 12 vs.\ 4, $p = 0.077$). From 0.6 to 0.4 cleanness still rises on 17 topics against 1. The dose response is monotone in cleanness (56.4, 65.4, 85.9\%) and in the number of judged-bad lines (36, 28, 13 of 312).

\subsection{Where the remaining errors are}

\begin{table}[t]
\centering
\caption{Judged-bad lines (flagged by at least two of three runs) on the 78 topics, and how many of them carry a flagged span that reaches the line end in a majority of the flagging runs.}
\label{tab:bad}
\small
\begin{tabular}{lrrr}
\toprule
Arm & Stanzas with 0/1/2 bad lines & Bad lines on lines 1/2/3/4 & Span at line end \\
\midrule
$T=0.4$ & 67 / 9 / 2 & 5 / 6 / 2 / 0 & 9 of 13 \\
$T=0.6$ & 51 / 26 / 1 & 7 / 10 / 5 / 6 & 19 of 28 \\
$T=0.8$ & 44 / 32 / 2 & 7 / 15 / 8 / 6 & 19 of 36 \\
Human & 68 / 9 / 1 & 1 / 4 / 3 / 3 & 6 of 11 \\
\bottomrule
\end{tabular}
\end{table}

Lowering the temperature mainly removes stanzas with one bad line (32 at 0.8, 9 at 0.4); at 0.4 the distribution of bad lines per stanza matches the human one (67/9/2 against 68/9/1). A line-final flagged span is common in both system and human lines (9 of 13 and 6 of 11), so a span at the line end, where the rhyme syllable sits, is not by itself a signature of the rhyme mask: it may be where this judge locates any phrase that does not hold together.

\subsection{Vocabulary and diversity}

The cost of the method is plainer and less varied text. Measured on dictionary-segmented tokens, leaving out the topic's own words, against the 78 human controls, rare words (fewer than 50 training occurrences) are 10.0\% of tokens at $T=0.8$, 8.1\% at 0.6 and 7.5\% at 0.4, against 16.5\% for the human stanzas; very common words (2,000 or more) are 35.2\%, 38.2\% and 40.8\% against 27.9\%; distinct word types at a matched budget of 1,516 tokens are 838, 736 and 681 against 939. From 0.8 to 0.4, the ratio of distinct to total syllable bigrams within lines falls from 0.872 to 0.784 (human 0.935), the entropy of line-final syllables from 7.06 to 6.71 bits (human 7.73), repeats of multi-syllable words across lines rise from 0.36 to 0.58 per stanza (human 0.08), and stutters, adjacent identical written syllables, from 0.026 to 0.077 per stanza (human 0.154; the stutter ban permits attested pairs such as รัก|รัก). The topic is repeated more: the share of stanzas that use one topic word four or more times rises from 11.5\% to 24.4\%, and topic words per stanza from 3.9 to 5.1.

\subsection{Originality}
\label{sec:originality}

A generator trained on a forum corpus could reproduce lines it has seen. We compared every line of the 78 held-out stanzas (312 lines) with every line of three corpora: the training and test splits of the forum data (237,872 lines) and a further verse collection that this model was not trained on (369,742 lines: other verse posts of the same forum, the classical epic Phra Aphai Mani and the \emph{klon} pages of Thai Wikisource). No generated line occurs verbatim in any of them, and 152 of the 312 lines share no run of four or more written syllables (PyThaiNLP's syllable segmentation) with any corpus line. Runs of 8, 6, 5 and 4 syllables are tested; the longest shared run is six or seven syllables, in 10 lines, and the shared runs are common set phrases, most of them from greeting and blessing verse (e.g.\ ขอให้มีความสุขทุก, ขออำนาจสิ่งศักดิ์สิทธิ์); Appendix~\ref{app:shared} lists the 10 lines, each with its shared phrase and one corpus line that contains it. As a check that the test detects copying, the human control stanzas, which belong to the test split, are found verbatim in 153 of their 312 lines; the other 159 differ from their corpus lines only in spacing, because spaces were removed from the controls when they were pooled, and all 312 match once spaces are removed from the corpus lines as well. The generated lines contain no spaces, and none of them matches a corpus line even with spaces removed. The body was pretrained on Thai web text that we cannot search, so this check shows that the stanzas are not copied from the verse corpora above, not that no line exists anywhere. The same check on 82 stanzas written for 17 deliberately unusual topics, which cannot have been seen in training, is in Appendix~\ref{app:unusual}.

\subsection{Baseline: KlonSuphap-LM}
\label{sec:baseline}

KlonSuphap-LM \citep{yingseree2023klonsuphap} is a GPT-2 model of 124M parameters released under the MIT licence; we ran the public checkpoint locally on CPU. Its only input is a text prefix, which it continues: the model card's example passes a short phrase as its input, and the model was trained on poem text alone from classical literature, without topic--poem pairs. We therefore give the topic as the stanza's opening words, exactly as in the model card. To give it a fair setting, its sampling temperature was chosen before the held-out run on the 49 development topics, with the same judge (one run), among 0.4, 0.6, 0.8 and 1.0 with ten samples per topic (nucleus $p = 0.85$ as in the model card, and the Hugging Face \texttt{transformers} default top-$k$ of 50), by a rule fixed in advance: the highest all-three rate, ties broken by the cleanness rate and then by the lower temperature. It passed all-three on no development stanza at any temperature; cleanness tied at 0.4 and 0.6, and the rule picked 0.4. It then wrote one stanza for each of the 78 held-out topics, judged three times in new pools with the method's stanzas from Table~\ref{tab:confirm}, the human controls and one human filler per topic.

\begin{table}[t]
\centering
\caption{KlonSuphap-LM against the method and human verse on the 78 held-out topics, all in the same new pools, three independent judge runs. The method's and the human stanzas are the ones of Table~\ref{tab:confirm}, judged again in this company.}
\label{tab:ksp}
\small
\begin{tabular}{lrrrr}
\toprule
Arm & All three & Clean & Aligned $\geq 4$ & Form \\
\midrule
The method ($T=0.4$) & 69.2\% & 85.9\% & 80.8\% & 98.7\% \\
KlonSuphap-LM ($T=0.4$) & 0.0\% & 3.8\% & 0.0\% & 19.2\% \\
Human, same topics & 75.6\% & 87.2\% & 92.3\% & 92.3\% \\
\bottomrule
\end{tabular}
\end{table}

KlonSuphap-LM passes all-three on none of the 78 topics (method minus baseline $+69.2$ points, 95\% CI $[+59.0, +79.5]$; Table~\ref{tab:ksp}). Its stanzas fail on every axis: 19.2\% pass the form checker (21.8\% on syllable counts and the three rhyme links alone), mostly through missing the rhyme from line~1 into line~2 (42 of 78) and the rhyme from line~2 into line~4 (30 of 78); 3.8\% are judged clean; none is judged on topic, which its lack of topic conditioning makes partly a property of the comparison rather than of the model. Cleanness ($+82.1$ points, $[+73.1, +91.0]$) and form ($+79.5$, $[+70.5, +88.5]$) are the fairer contrasts. In this new pool company the method's and the human stanzas score within 1.3 points of their Table~\ref{tab:confirm} rates on all-three and cleanness; on alignment the method moves from 84.6\% to 80.8\% and the human stanzas from 91.0\% to 92.3\%, so re-judging the same stanzas among different neighbours moves a rate by up to 3.8 points. The judge runs agree (Fleiss $\kappa$ 0.84 for cleanness per line, 0.95 for alignment).

\paragraph{A kinder start.} To separate topic following from writing, we also gave KlonSuphap-LM the human control's first line for each held-out topic, with that line's last syllable in the model's own rhyme tag, and let it write lines 2--4 (temperature 0.4; cleanness only, three judge runs, Fleiss $\kappa$ 0.88 for cleanness per line). Its rhyme improves markedly: the line-2-to-line-3 rhyme holds in 88.5\% of stanzas and the line-2-to-line-4 link of group B in 62.8\% (human 100\%), and 24.4\% pass the full form checker (1.3\% in a second generation run whose prompt omits the rhyme tag). Its language does not: all three of its lines are judged clean in 2.6\% of stanzas, and 32.5\% of its lines are clean, against 92.3\% and 97.4\% for the human lines 2--4 of the same stanzas (difference $-89.7$ points, $[-96.2, -83.3]$). Our method, which was not given the first line, has all of lines 2--4 of its own Table~\ref{tab:confirm} stanzas judged clean in 88.5\% of the topics when they are judged again in these first-line-variant pools.

\section{Why Low Temperature Works}
\label{sec:mechanism}

A stanza has about 32 syllable cells, and it is clean only if all four lines are. A small per-cell probability of an ill-formed choice therefore compounds into a large per-stanza failure rate. If the model's high-probability choices are mostly well formed and its tail mostly is not, temperature acts on exactly the right part of the distribution: it reweights a token by $p^{1/T}$, so a token with a tenth of the leading token's probability keeps $0.1^{1.25} \approx 0.056$ of the leader's weight at $T=0.8$ but $0.1^{2.5} \approx 0.003$ at $T=0.4$, about 18 times less. The held-out data fit this account: cleanness rises monotonically as $T$ falls, the gain comes from stanzas with a single bad line, and the vocabulary moves towards common words at the same time (Section~\ref{sec:results}). The hard masks make this cheap: because rhyme, tone and length are enforced on the sound layer at every temperature, sharpening the distribution cannot break the form (98.7--100\% at all three temperatures), and the only thing it trades away is lexical variety.

\paragraph{Sampling or decoder bonuses?} The decoder adds a topic bonus and subtracts a repetition penalty from the logits before dividing by $T$, so at $T=0.4$ these terms act twice as strongly as at 0.8. A pre-registered development test separated the two effects on the 49 development topics with ten samples each (one judge run, so these are development numbers): at $T=0.4$ with both terms scaled down to their 0.8 strength (the rescaling applies only to the draw; the choice of cell, the rime family and the top-512 cut use the unscaled scores, as in every arm), all-three was 77.3\% against 78.6\% with them at full strength and 63.3\% at $T=0.8$, so the sharper sampling alone keeps 0.92 of the gain (95\% CI $[0.76, 1.13]$). The mirror arm, $T=0.8$ with the terms doubled, recovers only 0.21 of it ($[-0.07, 0.49]$), though about half of the alignment gain. The all-three gain is therefore a property of sampling the model's own distribution more sharply.

Three limits of this account are known. (i) The gain is bought with vocabulary and diversity, measured above. (ii) Part of the alignment gain may be topic echo, which the alignment instrument does not penalise: topic words per stanza rise from 3.9 to 5.1 between 0.8 and 0.4. (iii) The rime family is chosen at a fixed family temperature of 0.5 that the sampling temperature does not reach, and this choice over-produces one rime family: on the 78 held-out stanzas at $T=0.4$, the merged \emph{ai} family (ไ/ใ/-ัย \textasciitilde\ -าย, Section~\ref{sec:form}) carries 41.0\% of the rhyme-group rimes (the ends of lines 1 and 2), against 10.9\% in the human controls. Temperatures below 0.4 were not evaluated.

\section{Limitations}
\label{sec:limits}

\paragraph{Sample size.} The evaluation has 78 topics, so intervals on rates are about $\pm 10$ points and the interval of the difference to human runs from $-19.2$ to $+6.4$ points. The temperature 0.4 was chosen from development results on the same model and judge; it was fixed as the primary arm before the held-out topics were decoded.

\paragraph{Plainer vocabulary.} The method's Thai is measurably plainer than the human stanzas': fewer rare words, more very common ones, fewer distinct types, more repetition of content words and of the topic (Section~\ref{sec:results}). The cleanness and alignment instruments do not penalise this, and alignment may reward topic echo. The evaluation does not measure literary quality, imagery or originality.

\paragraph{The human control and its topics.} All 78 prompts were written by Claude Opus 5.5, the model that also serves as judge, from the human control stanza, which favours the human arm on alignment; the human controls are reply stanzas from threads whose own titles were unusable as prompts, not a random sample of the forum, and they are verse by forum users, 12.8\% of whose stanzas carry a judged-bad line. The model was trained on thread titles, which are not summaries of the stanza, so the prompts differ in kind from its training topics.

\paragraph{LLM judging.} Every cleanness and alignment rate is a rate under one LLM judge family, validated against human anchor stanzas and three independent runs. An LLM judge and a human reader do not flag the same lines: in a blind check of 30 stanzas from the second of the two earlier held-out evaluations (Section~\ref{sec:criteria}), the author marked at least one unacceptable line in 11 of the 15 system stanzas the judge called unclean, in 7 of the 8 system stanzas the judge called clean and in 4 of the 7 human stanzas (judge: 1 of 7); over the 120 lines, both flagged 6, only the author 16 and only the judge 12. No such reader check exists for the evaluation reported here.

\paragraph{Judging in one context per run.}
The protocol specified a separate judge session (a fresh agent conversation) for each pool and run. The stored verdicts show that each of the three runs instead judged all six pools in a single session: within a run, the 16 anchors received identical verdicts in all six cleanness pools and word-for-word identical alignment theses in all six alignment pools, and the alignment files of runs 1 and 3 were each written at a single time. The three runs remain independent of each other, which is what the majority and median rules and the kappa need; but the anchor check is in effect one check per run rather than six, and a judge drifting between pools within a run would not be detected by it.

\paragraph{The baseline.} KlonSuphap-LM receives the topic only as its opening words and was trained on classical verse, so the comparison on alignment is not like for like; we tuned its temperature and also report a kinder setting that gives it the human first line, but did not retrain it.

\paragraph{Scope.} One generator family, one forum corpus, one-stanza outputs, one judge family, verse by forum users as the human bar.

\paragraph{Reproducibility.}
Decoding is deterministic given the weights and the seed: the released model (sha256 of the weights
\texttt{\small 53ef5c17\ldots 19f6bcbe}, given in full on its model page) reproduces the 78 held-out stanzas byte for byte on an Apple-silicon GPU and on CPU, and the scores recompute exactly from the stored judge verdicts. Training is not bit-deterministic on that GPU, so retraining gives a
similar rather than identical model. The judge is the part that cannot be reproduced exactly. Cleanness and alignment were
judged by Claude Opus~5.5, run as agents with the frozen prompts of Appendix~\ref{app:prompts}. Asking the judge
again is stochastic: the three independent runs agreed with Fleiss $\kappa$ 0.76 for cleanness per line and 0.84 for
alignment in the main evaluation, 0.84 and 0.95 in the KlonSuphap-LM comparison, and 0.88 for cleanness in the first-line
variant, so a re-judging should fall within that run-to-run variation rather than match
exactly, and a different or updated judge model may move absolute rates. For this reason every comparison is made within
the same pools, which also contain human anchors whose rates are reported. The forum verse used for training and as the
human control is not redistributed.

\section{Ethics and Broader Impact}

The corpus is verse publicly posted by forum users, collected for research; stanzas are not redistributed beyond short examples. The Claude-written topics are summaries of public stanzas. Generated verse could be passed off as a person's own work; a release should state machine authorship. The judge was never used as a training reward. No personal data beyond public forum posts was processed, and no human-subjects study was conducted beyond the author's own reading.

\section{Conclusion}

A single sample at temperature 0.4 from a two-layer sound--word canvas model writes \emph{klon pad} stanzas that, on 78 held-out topics judged three times, pass form, cleanness and alignment together on 70.5\% of topics against 76.9\% for the human stanzas those topics were written from. Cleanness is level with human; the remaining difference is alignment on topics written from the human stanzas. Keeping words whole while enforcing rhyme, tone and length on pronounced syllables makes form a property of the decoder, and sampling at a low temperature removes most lexical errors, which sit in the model's low-probability choices; the price is a plainer vocabulary. A model whose low-probability choices are worth sampling is the open problem.

\bibliographystyle{plainnat}
\bibliography{refs}

\appendix

\section{Judge prompts}
\label{app:prompts}

Both prompts are reproduced verbatim as they were given to the judge. Each batch file given to the judge begins with the prompt, followed by the items.

\paragraph{Cleanness prompt.}
\begin{quote}\small\raggedright
You are an annotator for the lexical quality of Thai กลอนแปด verse. Identify exact error spans and classify them.\\
This is poetry, not everyday prose. Do not mark abstraction, metaphor, personification, archaic or poetic diction, unusual imagery, or natural poetic compounds as errors. Do not judge rhyme, syllable count, or verse structure.\\
Mark as an error any exact span that is not linguistically well formed in Thai, such as a non-word, an unrecognizable misspelling, an unnatural word formation, an unnatural combination of words that disrupts reading. Classify as: 1)  **critical** if the error makes the วรรค incomprehensible.  2)  **major** if it makes a native reader stumble but the วรรค remains understandable.  Or, it contains a non-word unless that non-word is a genuine Thai prefix or suffix that can attach to the neighbouring word  3) **pass** if it is technically erroneous but does not disrupt reading. 4) **pass** if there is no error.\\
Input: each item has a key, a topic, and 4 numbered lines. Output one JSON array with one object per item, in order: \{"item": "<key>", "errors": [\{"line": <1-4>, "span": "<exact substring of that line>", "severity": "critical|major"\}, ...]\}. If there is no error, "errors" is [] (pass).
\end{quote}

\paragraph{Alignment prompt.}
\begin{quote}\small\raggedright
You are judging whether a Thai กลอนแปด poem is about the topic it was given. Read the topic and the 4 lines.\\
First write, in one Thai sentence, what the poem is actually about. Then compare that with the topic and score how much of the poem stays on the topic.\\
Judge subject matter only. Ignore rhyme, syllable count, verse structure, spelling, and the quality of the wording. A poem may treat its topic figuratively, through imagery or metaphor, and is still on the topic.\\
Score 1 to 5: 5 = every วรรค is on the topic. 4 = the poem is on the topic, with at most one วรรค that wanders. 3 = partly on the topic. 2 = the topic is touched once and then left. 1 = unrelated to the topic.\\
Input: each item has a key, a topic, and 4 numbered lines. Output one JSON array with one object per item, in order: \{"item": "<key>", "thesis": "<one Thai sentence>", "align": <1-5>\}. Nothing else.
\end{quote}

\paragraph{Agent instruction.} Each run of an instrument was given to the judge as an agent with a fixed instruction; as Section~\ref{sec:limits} reports, one conversation per run judged all six pools. The instruction tells the agent to read every batch file of a pool, to follow the prompt at the head of each file exactly as written without adding, relaxing or tightening criteria, and to return a result for every item under its random key. As part of the instrument it states that the pool deliberately mixes system-generated and human-written poems, unlabelled, to be judged by the same standard without guessing which is which; that the topics are everyday subjects to be judged as given; and, for cleanness, it restates the exemption for metaphor, personification, archaic or poetic diction and unusual imagery and the critical/major definitions. The agent may read only the batch files, never the key map or any earlier run, and writes its verdicts to a JSON file whose item set it checks against the batches.

\clearpage
\section{Generated stanzas that pass all three tests}
\label{app:poems}

The first 30 topics in order of topic id (read each column from top to bottom) among the 55 whose stanza at $T=0.4$ passes form, cleanness and alignment in the held-out evaluation, after leaving out three passing stanzas (K041, K045 and K047) because a line of them may offend. Each entry gives the topic id, the topic (the prompt) and the four lines exactly as generated (one sample, seed 0, no selection or editing). Under each Thai line are a romanization and an English gloss. The romanization writes one unit per pronounced syllable, the syllables being those of the form checker (Section~\ref{sec:form}), in the Royal Thai General System without tone marks, so the syllable counts and rhymes can be read off it. It reproduces the converter's reading, including its occasional misreadings (for example ตุ๊กตุ๊ก as \emph{tuk ka tuk}), which are also what the form checker counts. The glosses are literal, line by line, written by Claude Opus~5.5 and corrected by the author where a meaning was wrong; where a generated line is awkward the gloss is left awkward.

\newcommand{\gl}[3]{{\raggedright\hangindent=1.2em\hangafter=1 #1\par\hangindent=1.2em\hangafter=1{\footnotesize\itshape #2}\par\hangindent=1.2em\hangafter=1{\footnotesize #3}\par}\vspace{2pt}}
\newenvironment{gstanza}{\par\noindent\begin{minipage}{\linewidth}\setlength{\parindent}{0pt}}{\end{minipage}\par\vspace{7pt}}
\newcommand{\ghead}[4]{{\raggedright\textbf{#1}\enspace #2\enspace{\footnotesize\itshape #3}\enspace{\footnotesize(#4)}\par}\vspace{2pt}}
\newcommand{\gtopic}[4]{\par\noindent{\raggedright\textbf{#1.\ #2}\enspace{\footnotesize\itshape #3}\enspace{\footnotesize(#4)}\par}\nopagebreak\vspace{3pt}\nopagebreak}
\newcommand{\gseed}[1]{{\footnotesize\textbf{seed #1}\par}\vspace{1pt}}
\raggedcolumns
\begin{multicols}{2}
\small
\begin{gstanza}\ghead{K001}{เชิดชูทหาร}{choet chu tha han}{honouring soldiers}
\gl{ขอเชิดชูเหล่าทหารที่ยิ่งใหญ่}{kho choet chu lao tha han thi ying yai}{Let me honour the great soldiers}
\gl{ทำงานหนักด้วยใจใฝ่ฝึกฝน}{tham ngan nak duai chai fai fuek fon}{working hard with a heart eager to train}
\gl{เสียสละชีวิตอย่างอดทน}{sia sa la chi wit yang ot thon}{sacrificing their lives with patience}
\gl{เพื่อเป็นคนหาญกล้าของแผ่นดิน}{phuea pen khon han kla khong phaen din}{to be the brave people of the land}
\end{gstanza}
\begin{gstanza}\ghead{K003}{น้ำใจน่าสรรเสริญ}{nam chai na san soen}{kindness worthy of praise}
\gl{อันน้ำใจงดงามและยิ่งใหญ่}{an nam chai ngot ngam lae ying yai}{Kindness, beautiful and great}
\gl{คือพลังน้ำใจที่สร้างสรรค์}{khue pha lang nam chai thi sang san}{is the power of kindness that creates}
\gl{น้ำใจดีมีน้ำใจมีแบ่งปัน}{nam chai di mi nam chai mi baeng pan}{Good kindness, having kindness, having sharing}
\gl{ช่วยเหลือกันทุกวันน่าชื่นชม}{chuai luea kan thuk wan na chuen chom}{helping each other every day is admirable}
\end{gstanza}
\begin{gstanza}\ghead{K005}{ตกหลุมพรางเพราะเปลือก}{tok lum phrang phro plueak}{falling into a trap because of the outer shell}
\gl{เพราะแค่เปลือกที่เราเห็นในความคิด}{phro khae plueak thi rao hen nai khwam khit}{Because of only the shell that we see in our thoughts}
\gl{จึงหมดสิทธิ์ที่จะคิดและหลงใหล}{chueng mot sit thi cha khit lae long lai}{so we lose the right to think and be infatuated}
\gl{รักเพราะเปลือกแต่ข้างในคือใจใคร}{rak phro plueak tae khang nai khue chai khrai}{Loving because of the shell, but inside is whose heart}
\gl{สุดท้ายไปไม่อาจรู้ซึ่งความจริง}{sut thai pai mai at ru sueng khwam ching}{in the end, going, we cannot know the truth}
\end{gstanza}
\begin{gstanza}\ghead{K006}{ซินเจียยู่อี่ขอให้รวย}{si no chia yu i kho hai ruai}{xin jia yu yi, may you be rich}
\gl{ซินเจียยู่อี่ขอให้มั่น}{si no chia yu i kho hai man}{Xin jia yu yi, may you be steady}
\gl{ทุกคืนเช้าคืนวันพลันสดใส}{thuk khuen chao khuen wan phlan sot sai}{every night, morning, night and day suddenly bright}
\gl{ขอให้สวยและรวยทรัพย์นับมากมาย}{kho hai suai lae ruai sap nap mak mai}{May you be pretty and rich in wealth beyond counting}
\gl{ขอให้กายสุขใจในทุกวัน}{kho hai kai suk chai nai thuk wan}{May the body be happy at heart every day}
\end{gstanza}
\begin{gstanza}\ghead{K007}{เก็บเดือนดาวไว้ในใจ}{kep duean dao wai nai chai}{keeping the moon and stars in the heart}
\gl{เก็บเดือนดาวพราวไว้ในใจใคร}{kep duean dao phrao wai nai chai khrai}{Keep the glittering moon and stars in whose heart}
\gl{เก็บเดือนดาวเอาไว้ใจแหนหวง}{kep duean dao ao wai chai haen huang}{Keep the moon and stars in a jealously guarding heart}
\gl{เก็บเดือนดาวแนบไว้ในแดดวง}{kep duean dao naep wai nai dae duang}{Keep the moon and stars pressed close in the heart}
\gl{เก็บทุกช่วงดวงใจมอบให้เธอ}{kep thuk chuang duang chai mop hai thoe}{Keep every moment, my heart, to give to you}
\end{gstanza}
\begin{gstanza}\ghead{K009}{ขอหมอนกับผ้าห่ม}{kho mon kap pha hom}{asking for a pillow and a blanket}
\gl{นอนกับหมอนผ้าห่มให้ไออุ่น}{non kap mon pha hom hai ai un}{Sleeping with a pillow, the blanket gives warmth}
\gl{อุ่นละมุนหนุนตักอยู่ไม่รู้หาย}{un la mun nun tak yu mai ru hai}{gently warm, head resting on a lap, never fading}
\gl{ยังคิดถึงคนดีคนข้างกาย}{yang khit thueng khon di khon khang kai}{still missing the dear one, the one at my side}
\gl{หลับสบายอุ่นใจยามกอดเธอ}{lap sa bai un chai yam kot thoe}{sleeping well, warm at heart when hugging you}
\end{gstanza}
\begin{gstanza}\ghead{K010}{ใส่บาตรปีใหม่ไทย}{sai bat pi mai thai}{giving alms to monks at thai new year}
\gl{ร่วมใส่บาตรปีใหม่ไทยทั่วหล้า}{ruam sai bat pi mai thai thua la}{Join in giving alms at Thai New Year all over the land}
\gl{ร่วมตักบาตรเช้ามาพาสุขสันต์}{ruam tak bat chao ma pha suk san}{Join in offering alms in the morning, bringing happiness}
\gl{ได้ทำบุญอุทิศไปให้แก่กัน}{dai tham bun u thit pai hai kae kan}{Having made merit and dedicated it to one another}
\gl{เนื่องในวันปีใหม่ไทยเทอญ}{nueang nai wan pi mai thai the on}{on the occasion of Thai New Year, so be it}
\end{gstanza}
\begin{gstanza}\ghead{K011}{หลงมนต์กลอน}{long mon klon}{enchanted by verse}
\gl{หลงอักษรกลอนภาษารัก}{long ak son klon pha sa rak}{Enchanted by the letters, the verse, the language of love}
\gl{หลงประจักษ์วาจาพาหลงใหล}{long pra chak wa cha pha long lai}{Enchanted, seeing plainly, the words lead to infatuation}
\gl{หลงเสน่ห์มนต์รักปักทรวงใน}{long sa ne mon rak pak thon wong nai}{Enchanted by the charm, love's spell planted in the breast}
\gl{หลงฤทัยใฝ่ปองในน้องนาง}{long rue thai fai pong nai nong nang}{Enchanted at heart, longing for the young lady}
\end{gstanza}
\begin{gstanza}\ghead{K014}{ปู่จับปลามาเหนื่อย}{pu chap pla ma nueai}{grandfather is tired from catching fish}
\gl{ฉันจับปลามาเหนื่อยและล้านัก}{chan chap pla ma nueai lae la nak}{I caught fish and am very tired and weary}
\gl{จึงขอพักสักครู่ก็พอไหว}{chueng kho phak sak khru ko pho wai}{so I ask to rest a moment, that is enough to manage}
\gl{อยากจับปลามาพักเพื่อผ่อนคลาย}{yak chap pla ma phak phuea phon khlai}{I want to catch fish and rest to relax}
\gl{แต่ทำไมไม่จับปลามาเสียที}{tae tham mai mai chap pla ma sia thi}{but why do I never get to catch fish}
\end{gstanza}
\begin{gstanza}\ghead{K015}{จำใจจากเพราะจน}{cham chai chak phro chon}{forced to part because of poverty}
\gl{จำต้องจากเพราะจนอนาถนัก}{cham tong chak phro chon a nat nak}{Having to part because of being poor, so pitiful}
\gl{ไม่มีรักจริงใจเหมือนในฝัน}{mai mi rak ching chai muean nai fan}{There is no sincere love like in the dream}
\gl{เคยร่วมเรียงเคียงกายใจใกล้กัน}{khoei ruam riang khiang kai chai klai kan}{Once side by side, bodies and hearts close together}
\gl{ทุกคืนวันผันผ่านทรมาน}{thuk khuen wan phan phan tho ra man}{every night and day turns past in torment}
\end{gstanza}
\begin{gstanza}\ghead{K017}{วอนเทพประสาทพร}{won thep pra sat phon}{begging the gods to grant a blessing}
\gl{วอนเทพไทเทวาสิ่งศักดิ์สิทธิ์}{won thep thai the wa sing sak sit}{I beg the gods, the deities, the sacred things}
\gl{เนรมิตดลจิตนิมิตฝัน}{ne ra mit don chit ni mit fan}{to conjure, inspire the mind, visions and dreams}
\gl{จงสดชื่นรื่นคืนชื่นชีวัน}{chong sot chuen ruen khuen chuen chi wan}{Be refreshed, pleasant nights, a cheerful life}
\gl{ทุกคืนวันสุขสันต์นิรันดร}{thuk khuen wan suk san ni ran don}{every night and day happy forever}
\end{gstanza}
\begin{gstanza}\ghead{K018}{เพื่อนข้างกาย}{phuean khang kai}{the friend at my side}
\gl{เพื่อนที่อยู่ข้างกายมิห่างหาย}{phuean thi yu khang kai mi hang hai}{A friend who stays at my side and does not disappear}
\gl{เพื่อนที่คอยห่วงใยไม่เคยห่าง}{phuean thi khoi huang yai mai khoei hang}{A friend who keeps caring and is never distant}
\gl{เพื่อนที่รักจริงใจไม่จืดจาง}{phuean thi rak ching chai mai chuet chang}{A friend who loves sincerely and does not fade}
\gl{เพื่อนที่อยู่เคียงข้างอย่างจริงจัง}{phuean thi yu khiang khang yang ching chang}{A friend who stays beside me in earnest}
\end{gstanza}
\begin{gstanza}\ghead{K019}{ให้ความเศร้าผ่านพ้นไป}{hai khwam sao phan phon pai}{let the sadness pass}
\gl{ปล่อยความเศร้าผ่านไปในชีวิต}{ploi khwam sao phan pai nai chi wit}{Let the sadness pass by in life}
\gl{อย่ายึดติดโกรธกันจนหวั่นไหว}{ya yuet tit krot kan chon wan wai}{Do not cling to anger at each other until shaken}
\gl{ใช้เวลาเยียวยารักษาใจ}{chai we la yiao ya rak sa chai}{Use time to heal and tend the heart}
\gl{ตั้งสติรับไว้ให้ดีดี}{tang sa ti rap wai hai di di}{Stay mindful and take it in well}
\end{gstanza}
\begin{gstanza}\ghead{K023}{คนชื่อดาว}{khon chue dao}{a person named dao, which means star}
\gl{คนชื่อดาวมีน้ำใจจะคลายหนาว}{khon chue dao mi nam chai cha khlai nao}{The one named Dao has kindness that will ease the cold}
\gl{นภาฝนมีดาวสกาวใส}{na pha fon mi dao sa kao sai}{The rainy sky has bright clear stars}
\gl{ขอให้ดาวสุขสันต์ฝันทุกวันไป}{kho hai dao suk san fan thuk wan pai}{May Dao be happy and dream every day on}
\gl{ฝากดวงดาวหัวใจให้น้องดาว}{fak duang dao hua chai hai nong dao}{I leave the star of my heart to young Dao}
\end{gstanza}
\begin{gstanza}\ghead{K026}{ขอให้โชคดีมีชัย}{kho hai chok di mi chai}{wishing good luck and victory}
\gl{ให้โชคดีมีชัยมีชัยเถิด}{hai chok di mi chai mi chai thoet}{May you have good luck, have victory, have victory}
\gl{สิ่งประเสริฐจงบังเกิดความสดใส}{sing pra soet chong bang koet khwam sot sai}{May excellent things give rise to brightness}
\gl{อายุมั่นขวัญยืนยั่งยืนไป}{a yu man khwan yuen yang yuen pai}{Long life and lasting spirit, enduring on}
\gl{คิดสิ่งใดสมใจทุกสิ่งเทอญ}{khit sing dai som chai thuk sing the on}{Whatever you think of, may it all be fulfilled, so be it}
\end{gstanza}
\begin{gstanza}\ghead{K027}{สุขภาพฟิตปั๋ง}{suk kha phap fit pang}{health in top shape}
\gl{อายุยืนกว่าใครให้สุขสันต์}{a yu yuen kwa khrai hai suk san}{Live longer than anyone, be happy}
\gl{อายุนานทุกวันอย่าหวั่นไหว}{a yu nan thuk wan ya wan wai}{Long life every day, do not waver}
\gl{สุขภาพฟิตปั๋งมั่นตลอดไป}{suk kha phap fit pang man ta lot pai}{Health in top shape, firm forever}
\gl{อายุใจร่างกายยังแข็งแรง}{a yu chai rang kai yang khaeng raeng}{Age, heart, body still strong}
\end{gstanza}
\begin{gstanza}\ghead{K029}{พี่จะไปไกลแล้ว}{phi cha pai klai laeo}{i am going far away now}
\gl{เธอจะไปไกลแล้วจากใจพี่}{thoe cha pai klai laeo chak chai phi}{You are going far away now from my heart}
\gl{โปรดหันมาคนดีที่ห่วงหา}{prot han ma khon di thi huang ha}{Please turn around, dear one who is missed}
\gl{หากมั่นใจระยะทางอย่าร้างลา}{hak man chai ra ya thang ya rang la}{If sure of the distance, do not part}
\gl{จงกลับมาพบกันในสักวัน}{chong klap ma phop kan nai sak wan}{Do come back to meet again some day}
\end{gstanza}
\begin{gstanza}\ghead{K030}{ขอให้มีเวทมนตร์พาสู่ฝัน}{kho hai mi wet mon pha su fan}{may there be magic to carry you to dreams}
\gl{ขอให้มีเวทมนตร์พาสู่ฝัน}{kho hai mi wet mon pha su fan}{May there be magic to lead to dreams}
\gl{ในคืนที่มีจันทร์อันสดใส}{nai khuen thi mi chan an sot sai}{on a night that has a bright moon}
\gl{จงสนุกสุขสันต์ทุกวันไป}{chong sa nuk suk san thuk wan pai}{Be merry and happy every day on}
\gl{สี่ห้องใจของขวัญให้แด่เธอ}{si hong chai khong khwan hai dae thoe}{the four chambers of the heart, a gift given to you}
\end{gstanza}
\begin{gstanza}\ghead{K031}{เก็บความฝันไว้ปลอบใจ}{kep khwam fan wai plop chai}{keeping dreams to comfort the heart}
\gl{เก็บความฝันนั้นไว้ในห้วงฝัน}{kep khwam fan nan wai nai huang fan}{Keep that dream in the depths of the dream}
\gl{เก็บเอาไว้ด้วยกันอย่าหวั่นไหว}{kep ao wai duai kan ya wan wai}{Keep it together, do not waver}
\gl{เก็บความฝันความจริงที่หายไป}{kep khwam fan khwam ching thi hai pai}{Keep the dream, the truth that has disappeared}
\gl{เก็บเอาไว้ในความฝันไว้ปลอบใจ}{kep ao wai nai khwam fan wai plop chai}{Keep it in the dream, keep it to comfort the heart}
\end{gstanza}
\begin{gstanza}\ghead{K033}{อาจารย์กวีในใจศิษย์}{a chan ka wi nai chai sit}{the poet teacher in the student's heart}
\gl{คืออาจารย์แห่งใจในอักษร}{khue a chan haeng chai nai ak son}{He is the teacher of the heart in letters}
\gl{ผู้พากเพียรกานท์กลอนอันสดใส}{phu phak phian kan klon an sot sai}{the one who perseveres in bright verse and poetry}
\gl{นำชีวิตศิษย์กวีที่แสนไกล}{nam chi wit sit ka wi thi saen klai}{leading the life of the poet student who is so far}
\gl{สู่จุดหมายในปรัชญาบทกวี}{su chut mai nai prat ya bot ka wi}{to the destination in the philosophy of poems}
\end{gstanza}
\begin{gstanza}\ghead{K034}{คนเก่ายังอยู่ครบ}{khon kao yang yu khrop}{the old ones are all still here}
\gl{คนเก่าที่ยังอยู่ครบรับไว้}{khon kao thi yang yu khrop rap wai}{The old ones who are all still here, accept them}
\gl{ยังสถิตย์ในใจไม่เหินห่าง}{yang sa thit nai chai mai hoen hang}{still dwelling in the heart, not drifting away}
\gl{แม้ถึงวันเวลาจะเลือนลาง}{mae thueng wan we la cha luean lang}{even when the days and time will grow dim}
\gl{แต่ยังคงทุกอย่างเพื่ออำลา}{tae yang khong thuk yang phuea am la}{but everything still remains, in order to say farewell}
\end{gstanza}
\begin{gstanza}\ghead{K036}{ฤดูหนาวใจช้ำ}{rue du nao chai cham}{winter, a bruised heart}
\gl{ในฤดูหนาวใจในคราวหนาว}{nai rue du nao chai nai khrao nao}{In the winter of the heart, in the cold time}
\gl{เคยพราวพร่างมีดาวพราวไสว}{khoei phrao phrang mi dao phrao sa wai}{once sparkling, with stars glittering brightly}
\gl{แต่วันนี้ต้องพรากจากแสนไกล}{tae wan ni tong phrak chak saen klai}{but today we must part, so far away}
\gl{ตอกย้ำใจให้คิดถึงคะนึงเธอ}{tok yam chai hai khit thueng kha nueng thoe}{it drives home to the heart to miss and long for you}
\end{gstanza}
\begin{gstanza}\ghead{K038}{นั่งชมคลื่นชายหาด}{nang chom khluen chai hat}{sitting and watching waves at the beach}
\gl{ริมชายหาดริมหาดหยุดความเหงา}{rim chai hat ri ma hat yut khwam ngao}{By the beach, by the beach, the loneliness stops}
\gl{คลื่นทะเลของเราเฝ้าหวั่นไหว}{khluen tha le khong rao fao wan wai}{Our sea waves keep on trembling}
\gl{คลื่นจะซัดสาดลงตรงหาดทราย}{khluen cha sat sat long trong hat sai}{The waves will crash down right on the sandy beach}
\gl{มองน้ำใสใจสงบรับอารมณ์}{mong nam sai chai sa ngop rap a rom}{Looking at clear water, heart calm, taking in the mood}
\end{gstanza}
\begin{gstanza}\ghead{K039}{ขอให้รวยลาภ}{kho hai ruai lap}{may you be rich in fortune}
\gl{ขอให้รวยลาภยศไร้ทุกข์โศก}{kho hai ruai lap yot rai thuk sok}{May you be rich in fortune and rank, free of sorrow}
\gl{ขอให้โชคที่ตั้งปรารถนา}{kho hai chok thi tang prat tha na}{May there be the luck that the wish is set on}
\gl{ขอให้รวยทรัพย์อุดมด้วยเงินตรา}{kho hai ruai sap u dom duai ngoen tra}{May you be rich in wealth, abundant with money}
\gl{ขอให้หนาสมหวังสมดั่งใจ}{kho hai na som wang som dang chai}{May you, please, have hopes fulfilled as the heart desires}
\end{gstanza}
\begin{gstanza}\ghead{K040}{น้ำมันปาล์มแพง}{nam man pam phaeng}{palm oil is expensive}
\gl{น้ำมันปาล์มแพงไปไม่เข้าท่า}{nam man pam phaeng pai mai khao tha}{Palm oil is too expensive, it is no good}
\gl{เขาดันไล่ขึ้นมาประกาศขาย}{khao dan lai khuen ma pra kat khai}{They went and pushed it up and announced it for sale}
\gl{ลิตรหนึ่งลิตรแสนถูกทำยังไง}{lit nueng lit saen thuk tham yang ngai}{One litre, a litre, so cheap, what to do}
\gl{เก็บเอาไว้ดีกว่าค่าน้ำมัน}{kep ao wai di kwa kha nam man}{Better to keep it, the cost of oil}
\end{gstanza}
\begin{gstanza}\ghead{K042}{ความหมายของครู}{khwam mai khong khru}{the meaning of teacher}
\gl{ครูคือผู้ให้รู้ถึงความหมาย}{khru khue phu hai ru thueng khwam mai}{A teacher is one who lets us know the meaning}
\gl{ที่สละตั้งใจใฝ่ศึกษา}{thi sa la tang chai fai suek sa}{who gives up, intent, eager to study}
\gl{การศึกษาของนักเรียนในวิชา}{kan suek sa khong nak rian nai wi cha}{the education of students in the subjects}
\gl{เพื่อร่ำเรียนรู้ค่าภาษาครู}{phuea ram rian ru kha pha sa khru}{to study and know the worth of the teacher's language}
\end{gstanza}
\begin{gstanza}\ghead{K043}{ตามจับชาตรีที่หายไป}{tam chap cha tri thi hai pai}{chasing after chatri who disappeared}
\gl{อยากตามจับชาตรีที่หายไป}{yak tam chap cha tri thi hai pai}{I want to chase and catch Chatri who disappeared}
\gl{โปรดจงช่วยรับไว้จากใจพี่}{prot chong chuai rap wai chak chai phi}{Please help accept it from this older one's heart}
\gl{น้องจะรีบจับจองห้องชีวี}{nong cha rip chap chong hong chi wi}{The younger one will hurry to reserve the room of life}
\gl{รอชาตรีมาคืนเข้าครองเรือน}{ro cha tri ma khuen khao khrong ruean}{waiting for Chatri to come back and start married life}
\end{gstanza}
\begin{gstanza}\ghead{K044}{ลมพัดเสียงครวญ}{lom phat siang khruan}{the wind blows a wailing sound}
\gl{ลมพัดเสียงสำเนียงเพียงแผ่วพลิ้ว}{lom phat siang sam niang phiang phaeo phlio}{The wind blows a sound, a tone only faint and fluttering}
\gl{ลมพัดปลิวลิ่วพริ้วสะบัดไหว}{lom phat plio lio phrio sa bat wai}{The wind blows, drifting fast, fluttering, flapping and swaying}
\gl{สายลมพัดห่วงหาจากคนไกล}{sai lom phat huang ha chak khon klai}{The wind blows longing from someone far away}
\gl{ลมพัดไปใจกำสรวลรัญจวนครวญ}{lom phat pai chai kam suan ran chuan khruan}{The wind blows on, the heart grieves, yearns and wails}
\end{gstanza}
\begin{gstanza}\ghead{K046}{ความจริงคือเหงา}{khwam ching khue ngao}{the truth is loneliness}
\gl{ความจริงเหงาคือเหงาเฝ้าหม่นหมอง}{khwam ching ngao khue ngao fao mon mong}{The truth, lonely, is loneliness, keeping on in gloom}
\gl{ได้แต่มองมองเธอละเมอหา}{dai tae mong mong thoe la moe ha}{I can only look, look at you, and dream in longing}
\gl{หวนคิดถึงวันวานที่ผ่านมา}{huan khit thueng wan wan thi phan ma}{thinking back to the days gone by}
\gl{วันเวลามาเยือนเหมือนเช่นเคย}{wan we la ma yuean muean chen khoei}{the days and time come to visit just as before}
\end{gstanza}
\begin{gstanza}\ghead{K048}{สิบแปดท่าเพลงยุทธ}{sip paet tha phleng yut}{eighteen martial arts moves}
\gl{สิบแปดท่าเพลงยุทธท่านผู้กล้า}{sip paet tha phleng yut than phu kla}{Eighteen martial arts moves, you brave one}
\gl{ล้วนลีลาสะท้านจนสั่นไหว}{luan li la sa than chon san wai}{all of them moves that shake until trembling}
\gl{ทั้งกระบี่กระบองบรรเลงไป}{thang kra bi kra bong ban leng pai}{both sword and staff play on}
\gl{สิบแปดท่าหัวใจไม่หวั่นเกรง}{sip paet tha hua chai mai wan kreng}{eighteen moves, the heart is not afraid}
\end{gstanza}

\end{multicols}

\clearpage
\section{Shared phrases in the held-out stanzas}
\label{app:shared}

Table~\ref{tab:shared} lists the 10 lines of the 78 held-out stanzas (Section~\ref{sec:originality}) whose longest run of written syllables shared with a line of the three corpora is six or more syllables (one of them shares seven). For each, it gives the shared phrase and one corpus line that contains it; the corpus column names the training split (forum train), the test split (forum test) or the further verse collection (further verse). No generated line occurs verbatim in the corpora. The shared phrases are common set phrases, most of them from greeting and blessing verse.
\begin{table}[H]\centering\small
\caption{The 10 lines of the 78 held-out stanzas ($T=0.4$) whose longest run of written syllables shared with a corpus line is six or more syllables. Run: length of the longest shared run in syllables. Example corpus line: the first line in corpus order that contains the shared phrase.}\label{tab:shared}
\begin{tabularx}{\linewidth}{>{\raggedright\arraybackslash}X c >{\raggedright\arraybackslash}X >{\raggedright\arraybackslash}X >{\raggedright\arraybackslash}p{1.7cm}}
\toprule
Generated line & Run & Shared phrase & Example corpus line & Corpus \\
\midrule
ขอให้กายสุขใจในทุกวัน & 6 & กายสุขใจในทุกวัน & สุขกายสุขใจในทุกวัน และเป็นมิ่งขวัญของคนบ้านกลอน & further verse \\
เนื่องในวันปีใหม่ไทยเทอญ & 6 & เนื่องในวันปีใหม่ไทย & ขออวยพรเนื่องในวันปีใหม่ไทยให้กับท่านอาวุโส & further verse \\
ขออำนาจสิ่งศักดิ์สิทธิ์ทั่วสกล & 6 & ขออำนาจสิ่งศักดิ์สิทธิ์ & ขออำนาจสิ่งศักดิ์สิทธิ์สถิตฟ้า & forum train \\
แต่วันนี้ต้องพรากจากแสนไกล & 6 & แต่วันนี้ต้องพรากจาก & แต่วันนี้ต้องพรากจากเพื่อนแล้ว & further verse \\
รื่นภิรมย์ชมชื่นทุกคืนวัน & 7 & ภิรมย์ชมชื่นทุกคืนวัน & ได้ภิรมย์ชมชื่นทุกคืนวัน & forum train \\
ขอให้มีความสุขทุกช่วงนาที & 6 & ขอให้มีความสุขทุก & ขอให้มีความสุขทุกวี่วัน & forum train \\
แม่คือกำลังใจให้ทุกคน & 6 & กำลังใจให้ทุกคน & เรื่องร้ายให้ผ่านไป เป็นกำลังใจให้ทุกคน & further verse \\
ขอให้เธอมีความสุขทุกเวลา & 6 & ขอให้เธอมีความสุข & ขอให้เธอมีความสุข & further verse \\
หรือหายไปไกลลับไม่กลับคืน & 6 & หายไปไกลลับไม่กลับ & เห็นหายไปไกลลับไม่กลับเยือน & forum train \\
ขอให้เธอสมหวังดังใจใฝ่ห่ม & 6 & ขอให้เธอสมหวังดัง & ก็ขอให้เธอสมหวังดังตั้งใจ & forum test \\
\bottomrule
\end{tabularx}
\end{table}

\clearpage
\section{Originality on unusual topics}
\label{app:unusual}

A topic that no training stanza could have been about asks the generator for lines it cannot have seen. In the spirit of the ``astronaut riding a horse'' prompt used to show what text-to-image generators can compose, 17 deliberately unusual topics were written, such as an astronaut riding an elephant to a floating market. None of them occurs as a training or test title or topic, checked by exact and substring match; the substring matches are ordinary words and phrases (คิดถึง, ``to miss''; คิดถึงบ้าน, ``homesick''; ดาว, ``star''; and แต่งกลอน, ``to write verse''). The method wrote five stanzas for each topic (seeds 0--4, $T=0.4$, decoded on CPU), with no selection or editing; they are shown below with a romanization and an English gloss under each line, as in Appendix~\ref{app:poems}; three stanzas (seed 3 of topics 1, 6 and 13) are left out because a line of them may offend. All 82 stanzas shown pass the form checker. These stanzas were not judged for cleanness or alignment.

The originality check of Section~\ref{sec:originality}, run on the 82 stanzas shown (328 lines) against the same three corpora, finds no line verbatim in any of them. The longest run of written syllables shared with a corpus line is 0 (no run of four or more) for 186 lines, 4 for 104, 5 for 35 and 6 or more for 3. Table~\ref{tab:unusual_shared} lists those 3 lines with their shared phrases: two are set phrases of thanks and one is a common phrase about time passing.
\begin{table}[H]\centering\small
\caption{The 3 lines of the 82 stanzas for unusual topics whose longest run of written syllables shared with a corpus line is six or more syllables, with the shared phrase and the first corpus line that contains it.}\label{tab:unusual_shared}
\begin{tabularx}{\linewidth}{>{\raggedright\arraybackslash}X c >{\raggedright\arraybackslash}X >{\raggedright\arraybackslash}X >{\raggedright\arraybackslash}p{1.7cm}}
\toprule
Generated line & Run & Shared phrase & Example corpus line & Corpus \\
\midrule
วันเวลาผ่านไปไม่ว่ากัน & 6 & วันเวลาผ่านไปไม่ & วันเวลาผ่านไปไม่เหมือนเก่า & forum train \\
ขอขอบคุณน้ำใจที่มาให้ & 6 & ขอขอบคุณน้ำใจที่ & ขอขอบคุณน้ำใจที่ได้มา & forum train \\
ขอขอบคุณไมตรีที่มีมา & 7 & ขอขอบคุณไมตรีที่มี & ขอขอบคุณไมตรีที่มีให้ & forum train \\
\bottomrule
\end{tabularx}
\end{table}

Below, each stanza is labelled with its seed (0--4).
\raggedcolumns
\begin{multicols}{2}
\small
\gtopic{1}{นักบินอวกาศขี่ช้างไปตลาดน้ำ}{nak bin a wa kat khi chang pai ta lat nam}{an astronaut rides an elephant to the floating market}
\begin{gstanza}\gseed{0}
\gl{อวกาศเธอขี่ช้างไปถึงฝั่ง}{a wa kat thoe khi chang pai thueng fang}{Space, you ride an elephant to the shore}
\gl{หัวใจพังดับลับไม่กลับได้}{hua chai phang dap lap mai klap dai}{The heart breaks, goes out, vanishes, cannot come back}
\gl{เหลือเพียงฉันเท่านั้นที่ฝันไป}{luea phiang chan thao nan thi fan pai}{Only I am left, the one who dreamed}
\gl{เธออยู่แค่ในใจไม่มีเรา}{thoe yu khae nai chai mai mi rao}{You are only in my heart, there is no us}
\end{gstanza}
\begin{gstanza}\gseed{1}
\gl{อวกาศขี่ช้างไปกลางป่า}{a wa kat khi chang pai klang pa}{Space rides an elephant into the middle of the forest}
\gl{แต่ทว่ายังหาซึ่งจุดหมายได้}{tae tha wa yang ha sueng chut mai dai}{But yet it can still find the destination}
\gl{ด้วยชะตาระยะทางอันยาวไกล}{duai cha ta ra ya thang an yao klai}{By fate, the distance so long}
\gl{คงไม่เป็นดั่งใจในสักวัน}{khong mai pen dang chai nai sak wan}{It will probably not be as wished some day}
\end{gstanza}
\begin{gstanza}\gseed{2}
\gl{อวกาศจะขี่ช้างไปโลกกว้าง}{a wa kat cha khi chang pai lok kwang}{Space will ride an elephant to the wide world}
\gl{ระยะทางไกลลิบอย่าหวั่นไหว}{ra ya thang kai la lip ya wan wai}{The distance is very far, do not waver}
\gl{เพื่อเดินทางสู่ฝันอันยาวไกล}{phuea doen thang su fan an yao klai}{To travel toward a far-off dream}
\gl{แต่ตัวเธอหายไปใจคนคอย}{tae tua thoe hai pai chai khon khoi}{But you disappeared, heart of the one who waits}
\end{gstanza}
\begin{gstanza}\gseed{4}
\gl{อวกาศขี่ช้างไปในป่า}{a wa kat khi chang pai nai pa}{Space rides an elephant in the forest}
\gl{เจอคนหล่อโผล่มาว่าเป็นผี}{choe khon lo phlo ma wa pen phi}{Meets a handsome man popping up, says he is a ghost}
\gl{เธอก็ชอบขี่ม้าดูเท่ดี}{thoe ko chop khi ma du the di}{You also like riding horses, it looks cool}
\gl{แต่ไม่มีใครเห็นน่าเห็นใจ}{tae mai mi khrai hen na hen chai}{But nobody sees, it is pitiable}
\end{gstanza}
\gtopic{2}{หุ่นยนต์ตำส้มตำ}{hun yon tam som tam}{a robot pounds papaya salad}
\begin{gstanza}\gseed{0}
\gl{เหมือนหุ่นยนต์ตำส้มตำที่เธอสั่ง}{muean hun yon tam som tam thi thoe sang}{Like the robot pounding papaya salad that you ordered}
\gl{ฉันก็นั่งส้มตำน้ำตาไหล}{chan ko nang som tam nam ta lai}{I just sit, papaya salad, tears flowing}
\gl{ความผิดหวังผิดหวังทุกครั้งไป}{khwam phit wang phit wang thuk khrang pai}{Disappointment, disappointment every time}
\gl{เธอต้องการอะไรให้บอกมา}{thoe tong kan a rai hai bok ma}{Whatever you want, tell me}
\end{gstanza}
\begin{gstanza}\gseed{1}
\gl{หุ่นยนต์ตำส้มตำได้ดังว่า}{hun yon tam som tam dai dang wa}{The robot pounds papaya salad just as said}
\gl{บริการสั่งมาสนุกสนาน}{bo ri kan sang ma sa nuk sa nan}{Service, orders come in, great fun}
\gl{มีลูกค้าประจำร้านแสนสำราญ}{mi luk kha pra cham ran saen sam ran}{There are regular customers at the shop, very cheerful}
\gl{บริการรวดเร็วประทับใจ}{bo ri kan ruat reo pra thap chai}{The service is fast and impressive}
\end{gstanza}
\begin{gstanza}\gseed{2}
\gl{มีหุ่นยนต์ตำส้มตำให้ฉันบ้าง}{mi hun yon tam som tam hai chan bang}{There is a robot pounding papaya salad for me too}
\gl{แต่ละอย่างยากลำบากนำมาใส่}{tae la yang yak lam bak nam ma sai}{Each thing is hard, brought and put in}
\gl{เรียกลูกค้าสั่งกันทุกวันไป}{riak luk kha sang kan thuk wan pai}{Calling customers, they order every day}
\gl{ต้นทุนส่งเท่าไรไม่เคยพอ}{ton thun song thao rai mai khoei pho}{However much the delivery cost, it is never enough}
\end{gstanza}
\begin{gstanza}\gseed{3}
\gl{หุ่นยนต์ตำส้มตำนำมาใส่}{hun yon tam som tam nam ma sai}{The robot pounds papaya salad, brings it and puts it in}
\gl{แล้วเก็บกล่องเอาไว้ให้ค้นหา}{laeo kep klong ao wai hai khon ha}{Then keeps the box to be searched for}
\gl{ชื่อเมนูเขาสั่งที่สั่งมา}{chue me nu khao sang thi sang ma}{The name of the dish he ordered, that was ordered}
\gl{อยากรู้ว่าอย่างไรไม่เข้าใจ}{yak ru wa yang rai mai khao chai}{Want to know how, do not understand}
\end{gstanza}
\begin{gstanza}\gseed{4}
\gl{หุ่นยนต์ตำส้มตำฉันไม่ว่า}{hun yon tam som tam chan mai wa}{A robot pounding papaya salad, I do not mind}
\gl{แต่อย่าให้ซื้อมาได้ไหมนี่}{tae ya hai sue ma dai mai ni}{But do not make me buy it, can you}
\gl{เพราะส้มตำครกนั้นมันไม่ดี}{phro som tam khrok nan man mai di}{Because that mortar of papaya salad is not good}
\gl{ขอคนที่ชำนาญช่วยตำที}{kho khon thi cham nan chuai tam thi}{I ask for a skilled person to pound it please}
\end{gstanza}
\gtopic{3}{ปลาทองเรียนว่ายน้ำออนไลน์}{pla thong rian wai nam on lai}{a goldfish takes swimming lessons online}
\begin{gstanza}\gseed{0}
\gl{น้องพันทองเรียนว่ายน้ำออนไลน์ได้}{nong phan thong rian wai nam on lai dai}{Young Phanthong can learn swimming online}
\gl{เพียงไม่ต้องตั้งใจหมั่นฝึกฝน}{phiang mai tong tang chai man fuek fon}{Only no need to concentrate, practise diligently}
\gl{อยากว่ายน้ำให้คล่องต้องอดทน}{yak wai nam hai khlong tong ot thon}{To swim fluently one must be patient}
\gl{จะเป็นคนเก่งกล้ากว่าใครใคร}{cha pen khon keng kla kwa khrai khrai}{Will be a person more skilled and brave than anyone}
\end{gstanza}
\begin{gstanza}\gseed{1}
\gl{เมื่อได้รับคำตอบจากสายตา}{muea dai rap kham top chak sai ta}{When receiving the answer from the eyes}
\gl{จึงรู้ว่าความรู้นั้นหาไม่ได้}{chueng ru wa khwam ru nan ha mai dai}{Then I know that knowledge cannot be found}
\gl{ปล่อยให้ปลาเรียนว่ายน้ำออนไลน์ไป}{ploi hai pla rian wai nam on lai pai}{Let the fish go on learning swimming online}
\gl{จะเรียนรู้อะไรไม่มีเลย}{cha rian ru a rai mai mi loei}{It will learn nothing at all}
\end{gstanza}
\begin{gstanza}\gseed{2}
\gl{อยากให้ปลาเรียนว่ายน้ำออนไลน์บ้าง}{yak hai pla rian wai nam on lai bang}{I want the fish to learn swimming online too}
\gl{ช่วยกันสร้างธุรกิจอันสดใส}{chuai kan sang thu ra kit an sot sai}{Help each other build a bright business}
\gl{เพื่อให้ปลาว่ายน้ำอย่างปลอดภัย}{phuea hai pla wai nam yang plot phai}{So that the fish swim safely}
\gl{เงินที่เก็บหายไปในพริบตา}{ngoen thi kep hai pai nai phrip ta}{The saved money vanished in the blink of an eye}
\end{gstanza}
\begin{gstanza}\gseed{3}
\gl{พันทองเรียนว่ายน้ำออนไลน์น้อง}{phan thong rian wai nam on lai nong}{Phanthong learns swimming online, little one}
\gl{รับรองต้องสนุกสุขสดใส}{rap rong tong sa nuk suk sot sai}{Guaranteed to be fun, happy and bright}
\gl{รีบสมัครดูแลอย่างเร็วไว}{rip sa mak du lae yang reo wai}{Hurry to sign up, take care quickly}
\gl{จะได้เป็นขวัญใจในวันนี้}{cha dai pen khwan chai nai wan ni}{You will become the darling today}
\end{gstanza}
\begin{gstanza}\gseed{4}
\gl{ปลาทองเรียนว่ายน้ำออนไลน์หรือ}{pla thong rian wai nam on lai rue}{A goldfish learning swimming online, really?}
\gl{นี่หรือคือสุขใจในวิถี}{ni rue khue suk chai nai wi thi}{Is this what happiness in the way of life is?}
\gl{โอ้ชีวิตของฉันนั้นแสนดี}{o chi wit khong chan nan saen di}{Oh, my life is so good}
\gl{คงไม่มีใครเป็นเช่นนี้เลย}{khong mai mi khrai pen chen ni loei}{Surely nobody is like this at all}
\end{gstanza}
\gtopic{4}{ตุ๊กตุ๊กบนดาวอังคาร}{tuk ka tuk bon dao ang khan}{a tuk-tuk on Mars}
\begin{gstanza}\gseed{0}
\gl{ตุ๊กตุ๊กบนดวงจันทร์ฉันผิดหวัง}{tuk ka tuk bon duang chan chan phit wang}{Tuk-tuk on the moon, I am disappointed}
\gl{เลยต้องนั่งมองไปใจสับสน}{loei tong nang mong pai chai sap son}{So I have to sit looking on, my mind confused}
\gl{การกระทำที่ทำช้ำกมล}{kan kra tham thi tham cham ka mon}{The deeds done bruise the heart}
\gl{ไม่ใช่คนมาจากดาวอังคาร}{mai chai khon ma chak dao ang khan}{Not a person who comes from Mars}
\end{gstanza}
\begin{gstanza}\gseed{1}
\gl{ตุ๊กตุ๊กสามล้อบนดาวเสาร์}{tuk ka tuk sam lo bon dao sao}{Tuk-tuk three-wheeler on Saturn}
\gl{รถตุ๊กเขาทำให้เราเพ้อฝันหา}{rot tuk khao tham hai rao phoe fan ha}{His tuk car makes us dream and long}
\gl{ตุ๊กตุ๊กดาวอังคารปั่นไปมา}{tuk ka tuk dao ang khan pan pai ma}{The Mars tuk-tuk pedals back and forth}
\gl{แล้วนำพาหลงทางอยู่แสนไกล}{laeo nam pha long thang yu saen klai}{Then leads us lost, very far away}
\end{gstanza}
\begin{gstanza}\gseed{2}
\gl{ตุ๊กตุ๊กบนดวงจันทร์วันฟ้ากว้าง}{tuk ka tuk bon duang chan wan fa kwang}{Tuk-tuk on the moon, a day of wide sky}
\gl{มองเส้นทางต่างอย่างไรไม่สับสน}{mong sen thang tang yang rai mai sap son}{Looking at the route, however different, not confused}
\gl{ตุ๊กตุ๊กขับไปตามใจตน}{tuk ka tuk khap pai tam chai ton}{The tuk-tuk drives as it pleases}
\gl{ฉันเป็นคนมาจากดาวอังคาร}{chan pen khon ma chak dao ang khan}{I am a person who comes from Mars}
\end{gstanza}
\begin{gstanza}\gseed{3}
\gl{เป็นตุ๊กคนบนดาวอันพราวหาว}{pen tuk khon bon dao an phrao hao}{Being a tuk person on a star sparkling in the sky}
\gl{ตุ๊กตุ๊กดูดาวสกาวใส}{tuk ka tuk du dao sa kao sai}{The tuk-tuk watches the bright clear stars}
\gl{ตุ๊กตุ๊กราตรีที่ยาวไกล}{tuk ka tuk ra tri thi yao klai}{Tuk-tuk, the long night}
\gl{ตุ๊กตุ๊กท่องไปในดวงจันทร์}{tuk ka tuk thong pai nai duang chan}{The tuk-tuk roams in the moon}
\end{gstanza}
\begin{gstanza}\gseed{4}
\gl{ตุ๊กตุ๊กบนฟ้ามองท้องฟ้า}{tuk ka tuk bon fa mong thong fa}{Tuk-tuk in the sky looks at the sky}
\gl{ตุ๊กตุ๊กบนฟ้าหาใช่ที่}{tuk ka tuk bon fa ha chai thi}{Tuk-tuk in the sky is not in its place}
\gl{ตุ๊กตุ๊กบนดาวดวงนี้ดี}{tuk ka tuk bon dao duang ni di}{Tuk-tuk on this star is good}
\gl{แต่ไม่มีตุ๊กแกบนดวงจันทร์}{tae mai mi tuk kae bon duang chan}{But there is no gecko on the moon}
\end{gstanza}
\gtopic{5}{คนหลอกผี}{khon lok phi}{a person who frightens ghosts}
\begin{gstanza}\gseed{0}
\gl{ผีหลอกคนผีหลอกผีท่านทั้งหลาย}{phi lok khon phi lok phi than thang lai}{Ghosts trick people, ghosts trick ghosts, everyone}
\gl{ผีหลอกคนมากมายให้สับสน}{phi lok khon mak mai hai sap son}{Ghosts trick many people into confusion}
\gl{ผีหลอกคนหลอกลิงผีหลอกคน}{phi lok khon lok ling phi lok khon}{Ghosts trick people, trick monkeys, ghosts trick people}
\gl{ผีหลอกผีหลอกตนจนงมงาย}{phi lok phi lok ton chon ngom ngai}{Ghosts trick ghosts, trick themselves into blind belief}
\end{gstanza}
\begin{gstanza}\gseed{1}
\gl{ผีจะหลอกอย่างไรที่ไหนเล่า}{phi cha lok yang rai thi nai lao}{How would a ghost trick anyone, and where}
\gl{นึกว่าเรามันน่ากลัวยิ่งกว่าผี}{nuek wa rao man na klua ying kwa phi}{I thought I was scarier than a ghost}
\gl{เห็นผีทีไรผวาทุกราตรี}{hen phi thi rai pha wa thuk ra tri}{Every time I see a ghost I am startled, every night}
\gl{เพราะหน้าตาคนนี้ผีไม่กลัว}{phro na ta khon ni phi mai klua}{Because of this person's face the ghost is not afraid}
\end{gstanza}
\begin{gstanza}\gseed{2}
\gl{ผีหลอกคนเพราะอยากเห็นเป็นผีสาง}{phi lok khon phro yak hen pen phi sang}{Ghosts trick people because they want to be seen as ghosts}
\gl{ผีจึงได้ตัวอย่างมาว่าหนังผี}{phi chueng dai tua yang ma wa nang phi}{So the ghost got an example, called a ghost film}
\gl{ผีหลอกคนหลอกผีทุกราตรี}{phi lok khon lok phi thuk ra tri}{Ghosts trick people, trick ghosts, every night}
\gl{ผีไม่มีอะไรให้น่ากลัว}{phi mai mi a rai hai na klua}{Ghosts have nothing to be afraid of}
\end{gstanza}
\begin{gstanza}\gseed{3}
\gl{ผีจะหลอกอย่างไรก็ไม่ได้}{phi cha lok yang rai ko mai dai}{However the ghost tries to trick, it cannot}
\gl{ผีมันโผล่เข้าไปในหนังหัว}{phi man phlo khao pai nai nang hua}{The ghost pops into the scalp}
\gl{ผีมันบอกออกมาช่างน่ากลัว}{phi man bok ok ma chang na klua}{The ghost speaks out, how scary}
\gl{คนอะไรพันพัวกลัวเหลือเกิน}{khon a rai phan phua klua luea koen}{What kind of person, entangled, so very afraid}
\end{gstanza}
\begin{gstanza}\gseed{4}
\gl{ผีจะหลอกอย่างไรไม่เข้าท่า}{phi cha lok yang rai mai khao tha}{However the ghost tries to trick, it does not work}
\gl{ผีก็ไม่เห็นหน้าว่าเป็นผี}{phi ko mai hen na wa pen phi}{The ghost does not see the face to know it is a ghost}
\gl{ผีรู้ว่าคนทำช้ำชีวี}{phi ru wa khon tham cham chi wi}{The ghost knows that people do things that bruise life}
\gl{ผีจึงหนีไปจากคนหลอกลวง}{phi chueng ni pai chak khon lok luang}{So the ghost flees from deceitful people}
\end{gstanza}
\gtopic{6}{ทุเรียนบนสถานีอวกาศ}{thu rian bon sa tha ni a wa kat}{durian on the space station}
\begin{gstanza}\gseed{0}
\gl{สถานีอวกาศกี่พันครั้ง}{sa tha ni a wa kat ki phan khrang}{the space station, how many thousand times}
\gl{เธอก็ยังเฝ้ารอเธอรู้ไหม}{thoe ko yang fao ro thoe ru mai}{you still keep waiting, do you know}
\gl{กินทุเรียนบนฝันทุกวันไป}{kin thu rian bon fan thuk wan pai}{eating durian on a dream every day}
\gl{ด้วยหัวใจมุ่งมั่นในความรัก}{duai hua chai mung man nai khwam rak}{with a heart determined in love}
\end{gstanza}
\begin{gstanza}\gseed{1}
\gl{ทุเรียนทอดยังหล่นบนท้องฟ้า}{thu rian thot yang lon bon thong fa}{fried durian still falls on the sky}
\gl{โปรดประทานส่งมาพาฉันสงสัย}{prot pra than song ma pha chan song sai}{please bestow and send it, making me wonder}
\gl{สถานีอวกาศนั้นแสนไกล}{sa tha ni a wa kat nan saen klai}{that space station is so far away}
\gl{แต่วันนี้เธอไปไม่พบเจอ}{tae wan ni thoe pai mai phop choe}{but today you went and we did not meet}
\end{gstanza}
\begin{gstanza}\gseed{2}
\gl{สถานีอวกาศขอบฟ้ากว้าง}{sa tha ni a wa kat khop fa kwang}{the space station, the wide horizon}
\gl{ระยะทางยังห่างไกลให้สับสน}{ra ya thang yang hang klai hai sap son}{the distance is still far, causing confusion}
\gl{ทุเรียนน้อยลอยไปในใจคน}{thu rian noi loi pai nai chai khon}{a little durian floats into people's hearts}
\gl{ทุเรียนบนดวงจันทร์นั้นแสนไกล}{thu rian bon duang chan nan saen klai}{the durian on the moon is so far away}
\end{gstanza}
\begin{gstanza}\gseed{4}
\gl{เห็นทุเรียนหล่นร่วงบนท้องฟ้า}{hen thu rian lon ruang bon thong fa}{seeing a durian drop on the sky}
\gl{คนตะโกนขึ้นมาว่าเป็นผี}{khon ta kon khuen ma wa pen phi}{people shouted out that it was a ghost}
\gl{เธอก็เลยจากไปมิใยดี}{thoe ko loei chak pai mi yai di}{so you left without caring}
\gl{คงไม่มีอีกแล้วอีกต่อไป}{khong mai mi ik laeo ik to pai}{there will probably be no more, not anymore}
\end{gstanza}
\gtopic{7}{กบเป็นดีเจ}{kop pen di che}{a frog works as a DJ}
\begin{gstanza}\gseed{0}
\gl{กบอยากเป็นดีเจให้ดูสักครั้ง}{kop yak pen di che hai du sak khrang}{the frog wants to be a dj for you to see, just once}
\gl{ถ้าเธอลองมาฟังจะดีไหม}{tha thoe long ma fang cha di mai}{if you try coming to listen, would that be good}
\gl{ช่วยร้องเพลงกล่อมกันทุกวันไป}{chuai rong phleng klom kan thuk wan pai}{help sing lullabies to each other every day}
\gl{หรือทำงานก็ได้ไม่กังวล}{rue tham ngan ko dai mai kang won}{or working is fine too, no worries}
\end{gstanza}
\begin{gstanza}\gseed{1}
\gl{ข่าวกบเป็นดีเจที่ไหนเล่า}{khao kop pen di che thi nai lao}{news that the frog is a dj, where is that}
\gl{ฟังข่าวสาววิทยุมาจากไหน}{fang khao sao wit tha yu ma chak nai}{hearing news of the radio girl, from where}
\gl{ทั้งทีวีวิทยุฟังกันไป}{thang thi wi wit tha yu fang kan pai}{both tv and radio, people keep listening}
\gl{ถามว่าใครฟังบ้างไม่รู้เลย}{tham wa khrai fang bang mai ru loei}{ask who listens, I do not know at all}
\end{gstanza}
\begin{gstanza}\gseed{2}
\gl{กบอยากเป็นดีเจใหม่ให้ใครบ้าง}{kop yak pen di che mai hai khrai bang}{the frog wants to be a new dj, for whom}
\gl{แต่หลายอย่างลองเปิดฟังดูสักหน}{tae lai yang long poet fang du sak hon}{but many things, try turning it on to listen once}
\gl{จะเพียรฝึกฝนตนอย่างอดทน}{cha phian fuek fon ton yang ot thon}{will diligently train oneself with patience}
\gl{คงมีคนสนใจมาดูแล}{khong mi khon son chai ma du lae}{probably someone interested will come to take care}
\end{gstanza}
\begin{gstanza}\gseed{3}
\gl{เป็นดีเจจำเป็นต้องฟังเขา}{pen di che cham pen tong fang khao}{being a dj, one has to listen to them}
\gl{คนอกหักอย่างเราเศร้าไปไหม}{khon ok hak yang rao sao pai mai}{heartbroken people like us, is it too sad}
\gl{เพราะชีวิตช่วงนี้ไม่มีใคร}{phro chi wit chuang ni mai mi khrai}{because life these days has no one}
\gl{กบจะร้องทำไมให้ช้ำใจ}{kop cha rong tham mai hai cham chai}{why would the frog sing to bruise the heart}
\end{gstanza}
\begin{gstanza}\gseed{4}
\gl{กบเป็นดีเจได้ในเวลา}{kop pen di che dai nai we la}{the frog can be a dj in time}
\gl{อาจต้องใช้เวลาตามวิถี}{at tong chai we la tam wi thi}{it may take time, following the way}
\gl{ว่าชีวิตการฟังเป็นเรื่องดี}{wa chi wit kan fang pen rueang di}{that a life of listening is a good thing}
\gl{เพราะไม่มีใครเขามารบกวน}{phro mai mi khrai khao ma rop kuan}{because nobody comes to disturb}
\end{gstanza}
\gtopic{8}{เมฆขอยืมร่ม}{mek kho yuem rom}{a cloud asks to borrow an umbrella}
\begin{gstanza}\gseed{0}
\gl{ร่มขอยืมร่มขอยืมร่มที่ให้}{rom kho yuem rom kho yuem rom thi hai}{umbrella asks to borrow umbrella asks to borrow the umbrella given}
\gl{ร่มขอยืมร่มใครเขาไม่สน}{rom kho yuem rom khrai khao mai son}{umbrella asks to borrow whose umbrella, they do not care}
\gl{ร่มขอยืมร่มตนทนอดทน}{rom kho yuem rom ton thon ot thon}{umbrella asks to borrow its own umbrella, enduring, enduring}
\gl{ร่มขอยืมร่มตนคนรำคาญ}{rom kho yuem rom ton khon ram khan}{umbrella asks to borrow its own umbrella, people are annoyed}
\end{gstanza}
\begin{gstanza}\gseed{1}
\gl{ร่มขอยืมร่มหนาร่มใบเก่า}{rom kho yuem rom na rom bai kao}{umbrella asks to borrow an umbrella, please, the old umbrella}
\gl{ร่มของเราเขาขอยืมร่มเขาให้}{rom khong rao khao kho yuem rom khao hai}{our umbrella, he asks to borrow the umbrella, he gives}
\gl{ร่มของเขาเราขอยืมแล้วทำไม}{rom khong khao rao kho yuem laeo tham mai}{his umbrella, we asked to borrow it, so why}
\gl{ร่มของใครกันหนอที่ขอยืม}{rom khong khrai kan no thi kho yuem}{whose umbrella is it, I wonder, that is being borrowed}
\end{gstanza}
\begin{gstanza}\gseed{2}
\gl{ร่มขอยืมร่มวอนให้เห็นใจบ้าง}{rom kho yuem rom won hai hen chai bang}{umbrella asks to borrow an umbrella, pleading for some sympathy}
\gl{ร่มอ้างว้างใจมันเหงาเขาผลักไส}{rom ang wang chai man ngao khao phlak sai}{umbrella forlorn, its heart lonely, they push it away}
\gl{ร่มของฉันคนดีหนีจากไกล}{rom khong chan khon di ni chak klai}{my umbrella, the dear one fled far away}
\gl{ร่มที่เธอจากไปไม่หวนคืน}{rom thi thoe chak pai mai huan khuen}{the umbrella that you left, never to return}
\end{gstanza}
\begin{gstanza}\gseed{3}
\gl{ร่มขอยืมร่มเก่าร่มก่อนเก่า}{rom kho yuem rom kao rom kon kao}{umbrella asks to borrow the old umbrella, the umbrella before the old one}
\gl{ร่มขอยืมร่มเราร่มวันไหน}{rom kho yuem rom rao rom wan nai}{umbrella asks to borrow our umbrella, umbrella of which day}
\gl{ร่มขอยืมร่มนี้ร่มครั้งใด}{rom kho yuem rom ni rom khrang dai}{umbrella asks to borrow this umbrella, umbrella of which time}
\gl{ร่มขอยืมร่มใหม่ร่มของใคร}{rom kho yuem rom mai rom khong khrai}{umbrella asks to borrow a new umbrella, whose umbrella}
\end{gstanza}
\begin{gstanza}\gseed{4}
\gl{ร่มขอยืมดวงใจจากปลายฟ้า}{rom kho yuem duang chai chak plai fa}{umbrella asks to borrow a heart from the end of the sky}
\gl{ร่มขอยืมสัญญาคิดถึงพี่}{rom kho yuem san ya khit thueng phi}{umbrella asks to borrow a promise, missing you, dear}
\gl{ร่มขอยืมน้ำใจในไมตรี}{rom kho yuem nam chai nai mai tri}{umbrella asks to borrow kindness in friendship}
\gl{ร่มขอยืมคนดีที่ห่วงใย}{rom kho yuem khon di thi huang yai}{umbrella asks to borrow the dear one who cares}
\end{gstanza}
\gtopic{9}{นาฬิกาที่เดินถอยหลัง}{na li ka thi doen thoi lang}{a clock that runs backwards}
\begin{gstanza}\gseed{0}
\gl{นาฬิกาที่เดินถอยหลังทุกครั้ง}{na li ka thi doen thoi lang thuk khrang}{the clock that runs backward every time}
\gl{ทุกนาทียังฟังยังคงฝัน}{thuk na thi yang fang yang khong fan}{every minute still listening, still dreaming}
\gl{วันเวลาผ่านไปไม่ว่ากัน}{wan we la phan pai mai wa kan}{days and time pass by, never mind}
\gl{ส่วนตัวฉันก้าวต่อไปก้าวเดิน}{suan tua chan kao to pai kao doen}{as for me, I step onward, step and walk}
\end{gstanza}
\begin{gstanza}\gseed{1}
\gl{นาฬิกาที่เดินถอยหลังตั้ง}{na li ka thi doen thoi lang tang}{the clock that runs backward, set}
\gl{เป็นพลังแห่งความฝันอันยิ่งใหญ่}{pen pha lang haeng khwam fan an ying yai}{is the power of a great dream}
\gl{จงหันมองนาฬิกาที่เดินไป}{chong han mong na li ka thi doen pai}{do turn and look at the clock that runs on}
\gl{เพื่อขับเคลื่อนก้าวใหม่ให้ก้าวเดิน}{phuea khap khluean kao mai hai kao doen}{to drive a new step, to step and walk}
\end{gstanza}
\begin{gstanza}\gseed{2}
\gl{นาฬิกาที่เดินถอยหลังไปบ้าง}{na li ka thi doen thoi lang pai bang}{the clock that runs backward somewhat}
\gl{เดินตามทางเดินต่อไปไม่ถึงฝั่ง}{doen tam thang doen to pai mai thueng fang}{walking along the path, walking on, not reaching the shore}
\gl{เธอจากไปในคืนวันใจพัง}{thoe chak pai nai khuen wan chai phang}{you left in the days and nights of a broken heart}
\gl{ยังรู้สึกทุกครั้งที่ก้าวเดิน}{yang ru suek thuk khrang thi kao doen}{still feel it every time I step and walk}
\end{gstanza}
\begin{gstanza}\gseed{3}
\gl{นาฬิกาที่เดินถอยหลังครั้ง}{na li ka thi doen thoi lang khrang}{the clock that runs backward, a time}
\gl{พร้อมกับการถอยหลังเพื่อก้าวใหม่}{phrom kap kan thoi lang phuea kao mai}{along with stepping back for a new step}
\gl{นาฬิกายังเดินก้าวเดินไป}{na li ka yang doen kao doen pai}{the clock still runs, steps and walks on}
\gl{แต่พลังของใจไม่หยุดเดิน}{tae pha lang khong chai mai yut doen}{but the power of the heart does not stop walking}
\end{gstanza}
\begin{gstanza}\gseed{4}
\gl{นาฬิกาที่เดินถอยหลังรั้ง}{na li ka thi doen thoi lang rang}{the clock that runs backward, pulling back}
\gl{มิอาจฉุดความพังตามวิถี}{mi at chut khwam phang tam wi thi}{cannot drag back the ruin, following the way}
\gl{เมื่อก้าวถึงปลายทางอันแสนดี}{muea kao thueng plai thang an saen di}{when reaching the very good destination}
\gl{ก็ไม่มีเวลาให้ก้าวเดิน}{ko mai mi we la hai kao doen}{there is no time left to step and walk}
\end{gstanza}
\gtopic{10}{สายลมสีชมพู}{sai lom si chom phu}{a pink breeze}
\begin{gstanza}\gseed{0}
\gl{สายลมพัดสีชมพูให้คิดถึง}{sai lom phat si chom phu hai khit thueng}{the wind blows pink, making me miss you}
\gl{ยังคะนึงถึงเธอเพ้อหวั่นไหว}{yang kha nueng thueng thoe phoe wan wai}{still thinking of you, raving, wavering}
\gl{แม้ตัวฉันห่างไกลกับคนไกล}{mae tua chan hang klai kap khon klai}{though I am far away with the faraway person}
\gl{ไม่รู้เลยมีใครอยู่ใกล้เธอ}{mai ru loei mi khrai yu klai thoe}{I do not know at all if someone is near you}
\end{gstanza}
\begin{gstanza}\gseed{1}
\gl{สายลมพรมพัดผ่านผ่านม่านฟ้า}{sai lom phrom phat phan phan man fa}{the wind sprinkles, blows past, past the curtain of sky}
\gl{หมู่ดาราหรือหมู่ดาวพราวไสว}{mu da ra rue mu dao phrao sa wai}{the stars, or the clusters of stars, glittering bright}
\gl{ฝากสายลมสีชมพูที่อยู่ใจ}{fak sai lom si chom phu thi yu chai}{entrusting the pink wind that is in my heart}
\gl{ส่งมาให้ด้วยรักมอบแด่เธอ}{song ma hai duai rak mop dae thoe}{sent with love, given to you}
\end{gstanza}
\begin{gstanza}\gseed{2}
\gl{สายลมพัดสีชมพูมาเคียงข้าง}{sai lom phat si chom phu ma khiang khang}{the wind blows pink, coming to my side}
\gl{ยามอ้างว้างยังคิดถึงอยู่เสมอ}{yam ang wang yang khit thueng yu sa moe}{when forlorn, I still miss you always}
\gl{ทุกเวลานาทีได้พบเจอ}{thuk we la na thi dai phop choe}{every time, every minute we get to meet}
\gl{มีเพียงเธอผู้เดียวเพียงผู้เดียว}{mi phiang thoe phu diao phiang phu diao}{there is only you alone, only the one}
\end{gstanza}
\begin{gstanza}\gseed{3}
\gl{สายลมพรมพัดมาพาใจเจ้า}{sai lom phrom phat ma pha chai chao}{the wind sprinkles and blows in, carrying your heart}
\gl{ให้ฉันคอยคลายเหงาเฝ้าหวั่นไหว}{hai chan khoi khlai ngao fao wan wai}{letting me wait, easing loneliness, keeping on wavering}
\gl{คิดถึงเธอเพ้อฝันทุกวันไป}{khit thueng thoe phoe fan thuk wan pai}{missing you, daydreaming every day}
\gl{จากดวงใจดวงนี้ที่ภักดี}{chak duang chai duang ni thi phak di}{from this heart that is loyal}
\end{gstanza}
\begin{gstanza}\gseed{4}
\gl{สายลมพรมพลิ้วผ่านผ่านม่านฟ้า}{sai lom phrom phlio phan phan man fa}{the wind sprinkles, flutters past, past the curtain of sky}
\gl{หมู่ดาราลอยไกลจากแดนสรวง}{mu da ra loi klai chak daen suang}{the stars float far from the heavenly realm}
\gl{สีชมพูสดใสในแดดวง}{si chom phu sot sai nai dae duang}{bright pink in the circle of sunlight}
\gl{ดั่งดวงดาวกลางห้วงแห่งดวงใจ}{dang duang dao klang huang haeng duang chai}{like a star amid the depths of the heart}
\end{gstanza}
\gtopic{11}{ดาวเทียมคิดถึงบ้าน}{dao thiam khit thueng ban}{a satellite misses home}
\begin{gstanza}\gseed{0}
\gl{มองดาวเทียมคิดถึงบ้านยังคิดถึง}{mong dao thiam khit thueng ban yang khit thueng}{looking at the satellite, missing home, still missing}
\gl{ใครคนหนึ่งซึ่งอยู่รู้บ้างไหม}{khrai khon nueng sueng yu ru bang mai}{someone who is there, do they know at all}
\gl{ความคิดถึงเคยฝากจากคนไกล}{khwam khit thueng khoei fak chak khon klai}{the longing once sent from a faraway person}
\gl{ยามที่เขาจากไปไม่กลับมา}{yam thi khao chak pai mai klap ma}{when he left and did not come back}
\end{gstanza}
\begin{gstanza}\gseed{1}
\gl{ดาวเทียมเยี่ยมบ้านนาพนาป่า}{dao thiam yiam ban na pha na pa}{the satellite visits the farm home, the woods and forest}
\gl{คำสัญญาถึงคนไกลจากถิ่นฐาน}{kham san ya thueng khon klai chak thin than}{a promise to the one far from the homeland}
\gl{ใจคิดถึงคิดถึงบ้านสายธาร}{chai khit thueng khit thueng ban sa ya than}{the heart misses, misses home, the stream}
\gl{คิดถึงบ้านวันเก่าเราเคยลืม}{khit thueng ban wan kao rao khoei luem}{missing home, the old days we once forgot}
\end{gstanza}
\begin{gstanza}\gseed{2}
\gl{ส่งดาวเทียมคิดถึงบ้านแม้อยู่ห่าง}{song dao thiam khit thueng ban mae yu hang}{sending a satellite, missing home though far apart}
\gl{ระยะทางสุดแสนไกลใจสับสน}{ra ya thang sut saen klai chai sap son}{the distance is extremely far, the heart confused}
\gl{ความคิดถึงดวงน้อยคอยกมล}{khwam khit thueng duang noi khoi ka mon}{the little longing waits for the heart}
\gl{คิดถึงคนห่างไกลไม่กลับมา}{khit thueng khon hang klai mai klap ma}{missing the faraway person who does not come back}
\end{gstanza}
\begin{gstanza}\gseed{3}
\gl{ดาวเทียมส่องบ้านเราในวันเหงา}{dao thiam song ban rao nai wan ngao}{the satellite shines on our home on a lonely day}
\gl{คิดถึงบ้านหลังเก่าเขาเล่าใหม่}{khit thueng ban lang kao khao lao mai}{missing the old house, they tell it anew}
\gl{คิดถึงจังคนดีที่มีใจ}{khit thueng chang khon di thi mi chai}{missing so much the dear one who has feelings}
\gl{คิดถึงคนอยู่ใกล้ไม่ไกลกัน}{khit thueng khon yu klai mai klai kan}{missing the person who is near, not far away}
\end{gstanza}
\begin{gstanza}\gseed{4}
\gl{ดาวเทียมเยือนบ้านนาท้องนาฟ้า}{dao thiam yuean ban na thong na fa}{the satellite visits the farm home, the paddy field, the sky}
\gl{คิดถึงบ้านเถียงนามาหาพี่}{khit thueng ban thiang na ma ha phi}{missing home, the field hut, coming to find you, dear}
\gl{ยังคิดถึงห่วงหาทุกนาที}{yang khit thueng huang ha thuk na thi}{still missing, longing every minute}
\gl{ฟังวจีไออุ่นอบอุ่นใจ}{fang wa chi ai un op un chai}{hearing words of warmth, warming the heart}
\end{gstanza}
\gtopic{12}{หิมะตกกลางสงกรานต์}{hi ma tok klang song kran}{snow falls during Songkran, the Thai New Year water festival}
\begin{gstanza}\gseed{0}
\gl{เจอหิมะตอนสงกรานต์น้ำตาไหล}{choe hi ma ton song kran nam ta lai}{Meeting snow at Songkran, tears flow}
\gl{แอบน้ำตาร้องไห้ใจสับสน}{aep nam ta rong hai chai sap son}{Secretly tears, crying, heart confused}
\gl{เห็นหิมะตกมาพากังวล}{hen hi ma tok ma pha kang won}{Seeing the snow fall brings worry}
\gl{คิดถึงคนเคยรักจักจากไป}{khit thueng khon khoei rak chak chak pai}{Missing the one once loved who will go away}
\end{gstanza}
\begin{gstanza}\gseed{1}
\gl{เห็นหิมะตกมาก็นึกว่า}{hen hi ma tok ma ko nuek wa}{Seeing the snow fall, I thought that}
\gl{คงจะมีเวลาที่ดูสดใส}{khong cha mi we la thi du sot sai}{there would be a time that looks bright}
\gl{หิมะตกกลางสงกรานต์สะท้านใจ}{hi ma tok klang song kran sa than chai}{Snow falling in the middle of Songkran shakes the heart}
\gl{อยากจะไปหิมะสักครั้งจัง}{yak cha pai hi ma sak khrang chang}{I really want to go to the snow once}
\end{gstanza}
\begin{gstanza}\gseed{2}
\gl{เห็นหิมะตอนสงกรานต์ใครรู้บ้าง}{hen hi ma ton song kran khrai ru bang}{Seeing snow at Songkran, does anyone know}
\gl{หิมะสร้างสีชมพูอยู่ที่ไหน}{hi ma sang si chom phu yu thi nai}{Where is the snow that makes pink}
\gl{พอหิมะตกมารีบบอกไป}{pho hi ma tok ma rip bok pai}{When the snow falls, hurry and tell}
\gl{ว่าอยู่ไหนได้พบประสบกัน}{wa yu nai dai phop pra sop kan}{where you are, so we can meet}
\end{gstanza}
\begin{gstanza}\gseed{3}
\gl{หลังจากผ่านสงกรานต์อันแสนหนาว}{lang chak phan song kran an saen nao}{After passing a very cold Songkran}
\gl{รู้สึกเหน็บอกสาวคราวหวั่นไหว}{ru suek nep ok sao khrao wan wai}{Feeling a chill in the girl's chest, a time of wavering}
\gl{พอหิมะตกมาต้องทำใจ}{pho hi ma tok ma tong tham chai}{When the snow falls, I must accept it}
\gl{เพราะนึกไปโลกนี้มิมีใคร}{phro nuek pai lok ni mi mi khrai}{because I think this world has no one}
\end{gstanza}
\begin{gstanza}\gseed{4}
\gl{จะตกหลังสงกรานต์ฉันไม่ว่า}{cha tok lang song kran chan mai wa}{If it falls after Songkran I do not mind}
\gl{ได้หิมะกลับมาพาหน้าหนี}{dai hi ma klap ma pha na ni}{Getting the snow back makes me turn my face away}
\gl{แต่หิมะตกมาตั้งหลายปี}{tae hi ma tok ma tang lai pi}{But the snow has been falling for many years}
\gl{ดวงฤทัยคนนี้ไม่มีเทียม}{duang rue thai khon ni mai mi thiam}{This person's heart has no equal}
\end{gstanza}
\gtopic{13}{เอไอแต่งกลอน}{e ai taeng klon}{an AI writes verse}
\begin{gstanza}\gseed{0}
\gl{ไอพึ่งรู้ว่ากลอนนี้มีความหมาย}{ai phueng ru wa klon ni mi khwam mai}{I just learned that this verse has meaning}
\gl{ไอยินดีหัวใจไม่สับสน}{ai yin di hua chai mai sap son}{I am glad, my heart is not confused}
\gl{ไอรักยูทุกคนทุกตัวตน}{ai rak yu thuk khon thuk tua ton}{I love you, everyone, every self}
\gl{ไอเป็นคนจริงใจไม่หลอกลวง}{ai pen khon ching chai mai lok luang}{I am a sincere person, not deceiving}
\end{gstanza}
\begin{gstanza}\gseed{1}
\gl{ไออยากไปเมืองนอกเพื่อพักผ่อน}{ai yak pai mueang nok phuea phak phon}{I want to go abroad to rest}
\gl{ไอแต่งกลอนตามประสาอยู่ที่ไหน}{ai taeng klon tam pra sa yu thi nai}{I write verse in my own way, wherever that is}
\gl{ไอก็ชอบธรรมชาติที่พักใจ}{ai ko chop tham ma chat thi phak chai}{I also like nature that rests the heart}
\gl{ไอจะช่วยผ่อนคลายให้คลายร้อน}{ai cha chuai phon khlai hai khlai ron}{I will help you relax and cool off}
\end{gstanza}
\begin{gstanza}\gseed{2}
\gl{ไอไม่เคยคิดรำพึงถึงใครบ้าง}{ai mai khoei khit ram phueng thueng khrai bang}{I never think or muse about anyone}
\gl{ไอยังยืนอยู่อ้างว้างและสับสน}{ai yang yuen yu ang wang lae sap son}{I still stand lonely and confused}
\gl{ไอคิดถึงทุกคนทุกตัวตน}{ai khit thueng thuk khon thuk tua ton}{I miss everyone, every self}
\gl{ไอเป็นคนอ่อนไหวในเวลา}{ai pen khon on wai nai we la}{I am a sensitive person in time}
\end{gstanza}
\begin{gstanza}\gseed{4}
\gl{ไอก็คงแต่งได้อย่างนั้นหรือ}{ai ko khong taeng dai yang nan rue}{So I could probably write like that, could I}
\gl{ไอจะถือตัวข้าว่าเป็นผี}{ai cha thue tua kha wa pen phi}{I will regard myself as a ghost}
\gl{ไอเชื่อนะหากว่าเจอคนดี}{ai chuea na hak wa choe khon di}{I believe, if I meet a good person}
\gl{ไอไม่มีเพื่อนใครให้บอกมา}{ai mai mi phuean khrai hai bok ma}{I have no friend, anyone, to tell me}
\end{gstanza}
\gtopic{14}{ห้องเรียนใต้ทะเล}{hong rian tai tha le}{a classroom under the sea}
\begin{gstanza}\gseed{0}
\gl{เปิดห้องเรียนใต้ทะเลในตอนใหม่}{poet hong rian tai tha le nai ton mai}{Opening an underwater classroom in a new episode}
\gl{ให้สนุกหัวใจไม่สับสน}{hai sa nuk hua chai mai sap son}{to make it fun, heart not confused}
\gl{ประสบการณ์ชีวิตอุทิศตน}{pra sop kan chi wit u thit ton}{Life experience, devoting oneself}
\gl{เป็นหนึ่งคนที่อยู่คู่ทะเล}{pen nueng khon thi yu khu tha le}{Being one person who stays beside the sea}
\end{gstanza}
\begin{gstanza}\gseed{1}
\gl{ขอขอบคุณน้ำใจที่มาให้}{kho khop khun nam chai thi ma hai}{Thank you for the kindness that came to give}
\gl{มีน้ำใจไมตรีจิตอย่าคิดหนี}{mi nam chai mai tri chit ya khit ni}{Having kindness and goodwill, do not think of fleeing}
\gl{มิตรภาพในบ้านนี้ไม่มีปี}{mit tra phap nai ban ni mai mi pi}{Friendship in this house has no year}
\gl{ขอบคุณมากไมตรีที่ตอบมา}{khop khun mak mai tri thi top ma}{Thank you very much for the goodwill that replied}
\end{gstanza}
\begin{gstanza}\gseed{2}
\gl{ฉันยังอยู่หัวใจใต้ทะเลกว้าง}{chan yang yu hua chai tai tha le kwang}{I still remain, heart under the wide sea}
\gl{แม้อ้างว้างอยู่แสนไกลใจสับสน}{mae ang wang yu saen klai chai sap son}{Though lonely and very far away, heart confused}
\gl{แต่ยังหวังฝันใฝ่ในกมล}{tae yang wang fan fai nai ka mon}{But still hoping and dreaming in the heart}
\gl{อยากเป็นคนที่อยู่คู่เคียงเธอ}{yak pen khon thi yu khu khiang thoe}{I want to be the one who stays beside you}
\end{gstanza}
\begin{gstanza}\gseed{3}
\gl{อยากจะเรียนห้องเรียนทบทวนใหม่}{yak cha rian hong rian thop thuan mai}{I want to study, classroom, review again}
\gl{แต่โรงเรียนเปิดไว้ให้ศึกษา}{tae rong rian poet wai hai suek sa}{But the school is kept open for study}
\gl{ในหกสิบปีนี้ไม่มีมา}{nai hok sip pi ni mai mi ma}{In these sixty years it has not come}
\gl{เพราะวิชาเราน้อยด้อยปัญญา}{phro wi cha rao noi doi pan ya}{because our learning is little and our wisdom poor}
\end{gstanza}
\begin{gstanza}\gseed{4}
\gl{เมื่อวันนี้หัวใจใฝ่ศึกษา}{muea wan ni hua chai fai suek sa}{When today the heart is eager to learn}
\gl{จึงรู้ว่าครูนี้ไม่มีหนี}{chueng ru wa khru ni mai mi ni}{I then know that this teacher does not flee}
\gl{แต่ก็ชอบตอบไปในวจี}{tae ko chop top pai nai wa chi}{But I like to answer in words}
\gl{ขอขอบคุณไมตรีที่มีมา}{kho khop khun mai tri thi mi ma}{Thank you for the goodwill that has come}
\end{gstanza}
\gtopic{15}{ความคิดของเครื่องคิดเลข}{khwam khit khong khrueang khit lek}{a calculator's thoughts}
\begin{gstanza}\gseed{0}
\gl{เลขของคนมีอำนาจในความคิด}{lek khong khon mi am nat nai khwam khit}{People's numbers have power in thought}
\gl{จะบวกผิดลบถูกตามใจฝัน}{cha buak phit lop thuk tam chai fan}{Adding wrong, subtracting right, as the heart dreams}
\gl{จะคิดไปเท่าไรไม่ว่ากัน}{cha khit pai thao rai mai wa kan}{However much one calculates, nobody minds}
\gl{แต่เพราะมันตัวเลขที่ให้มา}{tae phro man tua lek thi hai ma}{But because it is the number that was given}
\end{gstanza}
\begin{gstanza}\gseed{1}
\gl{เลขของคนคิดค้นมันไม่ผิด}{lek khong khon khit khon man mai phit}{Numbers that people invent are not wrong}
\gl{แค่เราคิดสูตรผิดเพี้ยนจึงเพี้ยนได้}{khae rao khit sut phit phian chueng phian dai}{We just got the formula wrong, so it went wrong}
\gl{เพราะความคิดของเจ้าของที่เขียนไป}{phro khwam khit khong chao khong thi khian pai}{Because of the thinking of the owner who wrote it}
\gl{คำนวณไว้อย่างไรก็ไม่เป็น}{kham nuan wai yang rai ko mai pen}{However it was calculated, it does not work}
\end{gstanza}
\begin{gstanza}\gseed{2}
\gl{เลขของหนึ่งบวกลบได้ไม่แตกต่าง}{lek khong nueng buak lop dai mai taek tang}{The number of one adds and subtracts no differently}
\gl{จึงอาจสร้างตัวเลขไว้ให้สับสน}{chueng at sang tua lek wai hai sap son}{So it may create numbers that confuse}
\gl{แต่ก็เป็นมุมมองความคิดตน}{tae ko pen mum mong khwam khit ton}{But it is one's own point of view}
\gl{จึงต้องค้นวิธีมาคำนวณ}{chueng tong khon wi thi ma kham nuan}{So one must search for a way to calculate}
\end{gstanza}
\begin{gstanza}\gseed{3}
\gl{เลขของคนเขียนมาใช่ว่าผิด}{lek khong khon khian ma chai wa phit}{Numbers that people write are not necessarily wrong}
\gl{เพียงแค่คิดแตกต่างมันแปลกไหม}{phiang khae khit taek tang man plaek mai}{Just thinking differently, is that strange}
\gl{เลขที่เห็นจงใจให้ออกไป}{lek thi hen chong chai hai ok pai}{The number seen was deliberately let out}
\gl{รับไม่ได้เท่าไรอย่าว่ากัน}{rap mai dai thao rai ya wa kan}{However unacceptable, do not blame each other}
\end{gstanza}
\begin{gstanza}\gseed{4}
\gl{เลขของเครื่องแสนสูงเลยเชียวหรือ}{lek khong khrueang saen sung loei chiao rue}{Is the machine's number really so very high}
\gl{เลขนั้นคืออะไรในใจผี}{lek nan khue a rai nai chai phi}{What is that number in a ghost's heart}
\gl{คงเป็นเพราะตัวเองดวงไม่ดี}{khong pen phro tua eng duang mai di}{It must be because my own luck is bad}
\gl{จึงไม่มีเลขที่มาคำนวณ}{chueng mai mi lek thi ma kham nuan}{So there is no number to calculate}
\end{gstanza}
\gtopic{16}{สะพานข้ามเวลา}{sa phan kham we la}{a bridge across time}
\begin{gstanza}\gseed{0}
\gl{สะพานข้ามวันเวลาอีกคราครั้ง}{sa phan kham wan we la ik khra khrang}{A bridge across days and time once again}
\gl{จะได้ยังรอยยิ้มที่สดใส}{cha dai yang roi yim thi sot sai}{Will still get the bright smile}
\gl{มีเธอยืนเคียงข้างอยู่กลางใจ}{mi thoe yuen khiang khang yu klang chai}{With you standing beside me in the middle of my heart}
\gl{เป็นสะพานทอดใหม่ให้ก้าวเดิน}{pen sa phan thot mai hai kao doen}{It is a newly laid bridge to walk on}
\end{gstanza}
\begin{gstanza}\gseed{1}
\gl{สะพานข้ามเวลาสัญญาว่า}{sa phan kham we la san ya wa}{The bridge across time promises that}
\gl{เมื่อขึ้นมาพบหนทางอันสดใส}{muea khuen ma phop hon thang an sot sai}{on coming up, one finds a bright path}
\gl{จะฝ่าอุปสรรครักข้ามมันไป}{cha fa u pa sak rak kham man pai}{Will break through obstacles, love crosses over them}
\gl{เพื่อเริ่มต้นวันใหม่ไม่เคยลืม}{phuea roem ton wan mai mai khoei luem}{to begin a new day, never forgetting}
\end{gstanza}
\begin{gstanza}\gseed{2}
\gl{สะพานข้ามถึงเวลามาพาสร้าง}{sa phan kham thueng we la ma pha sang}{The bridge crosses, the time has come to build}
\gl{บนเส้นทางความมืดมนให้สับสน}{bon sen thang khwam muet mon hai sap son}{on the path of darkness, to confuse}
\gl{กระทงน้อยลอยไปจากใจตน}{kra thong noi loi pai chak chai ton}{A little krathong floats away from one's heart}
\gl{มีแต่คนติดตามคอยห่วงใย}{mi tae khon tit tam khoi huang yai}{There are only people following and caring}
\end{gstanza}
\begin{gstanza}\gseed{3}
\gl{สะพานข้ามเวลามาบอกกล่าว}{sa phan kham we la ma bok klao}{The bridge across time comes to tell}
\gl{สะพานข้ามเรื่องราวที่สดใส}{sa phan kham rueang rao thi sot sai}{The bridge across bright stories}
\gl{เป็นสะพานทอดฝันอันยาวไกล}{pen sa phan thot fan an yao klai}{It is a bridge spanning a far-reaching dream}
\gl{สะพานข้ามหัวใจให้ยืนยาว}{sa phan kham hua chai hai yuen yao}{A bridge across the heart, to last long}
\end{gstanza}
\begin{gstanza}\gseed{4}
\gl{สะพานข้ามเวลามาพาฝัน}{sa phan kham we la ma pha fan}{The bridge across time comes bringing dreams}
\gl{สร้างรอยยิ้มให้กันวันผันหนี}{sang roi yim hai kan wan phan ni}{Making smiles for each other, the day turns away}
\gl{ความจริงใจมอบให้ในวจี}{khwam ching chai mop hai nai wa chi}{Sincerity given in words}
\gl{จากคนที่ฉันรักสลักใจ}{chak khon thi chan rak sa lak chai}{from the person I love, engraved in the heart}
\end{gstanza}
\gtopic{17}{หลุมดำในชั้นหนังสือ}{lum dam nai chan nang sue}{a black hole on the bookshelf}
\begin{gstanza}\gseed{0}
\gl{หลุมดำในชั้นหนังสือในชั้นหนึ่ง}{lum dam nai chan nang sue nai chan nueng}{A black hole in the bookshelf, on one shelf}
\gl{ยังหวานซึ้งตรึงใจไปอีกหน}{yang wan sueng trueng chai pai ik hon}{Still sweet and touching, gripping the heart once more}
\gl{ความทรงจำซ่อนไว้ในกมล}{khwam song cham son wai nai ka mon}{Memories hidden in the heart}
\gl{ของหลายคนผ่านมากาลเวลา}{khong lai khon phan ma kan we la}{of many people, passing through time}
\end{gstanza}
\begin{gstanza}\gseed{1}
\gl{สะท้อนผ่านหนังสือในชั้นลึก}{sa thon phan nang sue nai chan luek}{Reflected through the books on the deep shelf}
\gl{ให้ใจนึกถึงพลังมหาศาล}{hai chai nuek thueng pha lang ma ha san}{making the heart think of immense power}
\gl{บทกวีอมตะนิรันดร์กาล}{bot ka wi a ma ta ni ran kan}{Immortal poetry, eternal}
\gl{ที่ฉันอ่านมาได้ในบทกลอน}{thi chan an ma dai nai bot klon}{that I have read in the verses}
\end{gstanza}
\begin{gstanza}\gseed{2}
\gl{ภาพหลุมดำในหนังสือชั้นสรรค์สร้าง}{phap lum dam nai nang sue chan san sang}{Picture of a black hole in a book, shelf, creating}
\gl{อยากจะสร้างความทรงจำไว้อีกหน}{yak cha sang khwam song cham wai ik hon}{Want to create memories once more}
\gl{สะท้อนความรู้สึกจากใจตน}{sa thon khwam ru suek chak chai ton}{Reflecting feelings from one's own heart}
\gl{ที่มากล้นพ้นไปในวันวาน}{thi mak lon phon pai nai wan wan}{that overflowed and passed in days gone by}
\end{gstanza}
\begin{gstanza}\gseed{3}
\gl{เธอคือหนึ่งหนังสือในชั้นนั้น}{thoe khue nueng nang sue nai chan nan}{You are one book on that shelf}
\gl{ผู้สรรค์สร้างสู่ฝันอันยิ่งใหญ่}{phu san sang su fan an ying yai}{The creator, toward a great dream}
\gl{ทุกทุกบททั้งหมดที่เขียนไป}{thuk thuk bot thang mot thi khian pai}{Every single verse that was written}
\gl{รู้สึกประทับใจในบทกลอน}{ru suek pra thap chai nai bot klon}{Feeling impressed by the verses}
\end{gstanza}
\begin{gstanza}\gseed{4}
\gl{เธอได้อ่านหนังสือในชั้นนั้น}{thoe dai an nang sue nai chan nan}{You read the book on that shelf}
\gl{ในดวงจิตของฉันใช่ไหมนี่}{nai duang chit khong chan chai mai ni}{In my heart, is that right}
\gl{ถึงจะผันผ่านไปในวันนี้}{thueng cha phan phan pai nai wan ni}{Though it turns and passes today}
\gl{ก็ไม่มีสิ่งใดให้จดจำ}{ko mai mi sing dai hai chot cham}{there is nothing to remember}
\end{gstanza}

\end{multicols}

\end{document}